\documentclass{article}

\usepackage[final]{neurips_2022}

\usepackage{wrapfig}
\usepackage{microtype}
\usepackage{graphicx}
\usepackage{subfigure}
\usepackage{floatrow}
\usepackage[]{algorithmic} 
\usepackage{booktabs} 
\usepackage{paralist}
\usepackage{xcolor}
\usepackage{times}
\usepackage{latexsym}
\usepackage[utf8]{inputenc}
\usepackage{times}
\usepackage{latexsym}
\usepackage[ruled,linesnumbered]{algorithm2e}

\newtheorem{propose}{Proposition}
\usepackage{url}
\usepackage{paralist}
\usepackage{multicol,graphicx,mathrsfs,pifont,amscd,latexsym,url, pifont,epstopdf,setspace,multirow,colortbl, bbm,listings, balance,indentfirst,enumitem,amsmath,amssymb, verbatim,microtype}
\usepackage{graphicx,booktabs, paralist}
\usepackage{blindtext}

\definecolor{darkblue}{rgb}{0.0, 0.2, 0.6}

\usepackage{color}
\usepackage[colorlinks,
linkcolor=black,
anchorcolor=darkblue,
citecolor=darkblue,
urlcolor=darkblue
]{hyperref}

\newcommand{\cmark}{\ding{51}}%
\newcommand{\xmark}{\ding{55}}%
\newfloatcommand{capbtabbox}{table}[][\FBwidth]

\title{Improving Multi-Task Generalization via Regularizing Spurious Correlation}

\author{%
   Ziniu Hu$^1$\thanks{This work was done when Ziniu was an intern at Google.}\
   , Zhe Zhao$^2$, Xinyang Yi$^2$, Tiansheng Yao$^2$, Lichan Hong$^2$, Yizhou Sun$^1$, Ed H. Chi$^2$  \\
      $^1$University of California, Los Angeles,\  \texttt{\{bull, yzsun\}@cs.ucla.edu}\\
      $^2$Google Research, Brain Team,\ \texttt{\{zhezhao,xinyang,tyao,lichan,edchi\}@google.com}
}

\begin{document}

\maketitle

\newcommand{\framework}{MT-CRL\xspace}
\newcommand{\method}{G-IRM}

\newcommand{\YS}[1]{{\bf\color{red}[{YS: #1}]}}

\begin{abstract}



Multi-Task Learning (MTL) is a powerful learning paradigm to improve generalization performance via knowledge sharing. However, existing studies find that MTL could sometimes hurt generalization, especially when two tasks are less correlated. One possible reason that hurts generalization is spurious correlation, i.e., some knowledge is spurious and not causally related to task labels, but the model could mistakenly utilize them and thus fail when such correlation changes. In MTL setup, there exist several unique challenges of spurious correlation. First, the risk of having non-causal knowledge is higher, as the shared MTL model needs to encode all knowledge from different tasks, and causal knowledge for one task could be potentially spurious to the other. Second, the confounder between task labels brings in a different type of spurious correlation to MTL. Given such label-label confounders, we theoretically and empirically show that MTL is prone to taking non-causal knowledge from other tasks. To solve this problem, we propose Multi-Task Causal Representation Learning (\framework) framework. \framework aims to represent multi-task knowledge via disentangled neural modules, and learn robust task-to-module routing graph weights
via MTL-specific invariant regularization. Experiments show that \framework could enhance MTL model's performance by 5.5$\%$ on average over Multi-MNIST, MovieLens, Taskonomy, CityScape, and NYUv2, and show it could indeed alleviate spurious correlation problem.

\end{abstract}

\section{Introduction}




Multi-Task Learning (MTL), a learning paradigm~\citep{caruana1997multitask, zhang2018overview} aiming to train a single model for multiple tasks, is expected to benefit the overall generalization performance than single-task learning~\citep{DBLP:journals/jmlr/MaurerPR16, DBLP:conf/nips/TripuraneniJJ20} given the assumption that there exists some common knowledge to handle different tasks.
However, recent studies observed that, when two tasks are less correlated, MTL could lead to even worse overall performance~\citep{DBLP:journals/corr/ParisottoBS15, DBLP:journals/corr/abs-2009-00909}. A line of works~\citep{DBLP:conf/nips/YuK0LHF20, DBLP:conf/iclr/WangTF021, DBLP:journals/corr/abs-2109-04617} resort performance drop to optimization challenge because conflicting tasks might compete for model capacity. 
However, both \citet{DBLP:conf/icml/StandleyZCGMS20} and our analysis in Section~\ref{sec:analysis} show that, even with an over-parameterized model that achieves low MTL training loss, the final generalization performance could be worse than single-task learning. This finding motivates us to think about the following question: Are there any intrinsic problems in MTL that hurt generalization?

One widely studied issue that influences generalization is the spurious correlation problem~\citep{DBLP:conf/iclr/GeirhosRMBWB19,DBLP:journals/natmi/GeirhosJMZBBW20}, i.e., correlation that only existed in training datasets due to unobserved confounders~\citep{LopezPaz2016FromDT}, but not causally correct. For example, as \citet{DBLP:conf/eccv/BeeryHP18} discussed, when we train an image classification model to identify cows with a biased dataset where cows mostly appear in pastures, the trained cow classification model could exploit the features of background (e.g., pastures) to make prediction. Thus, when we apply the classifier to another dataset where cows also appear in other locations such as farms or rivers, it will fail to generalize~\citep{DBLP:conf/iclr/NagarajanAN21}.


When it comes to MTL setting, there exist several unique challenges to handle spurious correlation problem. \textbf{First, the risk of having non-causal features is higher}. Suppose each task has different sets of causal features. To train a single model for all these tasks, the shared representation should encode all required features. Consequently, the causal features for one task could be potentially spurious to the other tasks, and such risk could be even higher with an increasing number of tasks. \textbf{Second, the confounder that leads to spurious correlation is different}. Instead of the standard confounders between feature and label, the nature of MTL brings in a unique type of confounders between task labels, e.g., correlation between tasks' labels could change in different distributions. 
For example, when we train a MTL model to solve both cow classification and scene recognition tasks, its encoder needs to capture both foreground and background information, and the spurious correlation between the two tasks in training set could mislead per-task model to utilize irrelevant information, e.g., use background to predict cow. 
Given such label-label confounders that are unique for MTL, we theoretically prove that MTL is prone to taking non-causal knowledge learned from other tasks. We then conduct empirical analysis to validate the hypothesis. In summary, we point out the unique challenges of spurious correlation in MTL setup, and show that it indeed influences multi-task generalization.



In light of the analysis, we try to solve the spurious correlation problem in MTL.
Among all the knowledge learned in the shared representation layer through end-to-end training, an ideal MTL framework should learn to leverage only the causal knowledge to solve each task by identifying the correct causal structure.
Following the recent advances that enable causal learning in an end-to-end learning model~\citep{DBLP:journals/pieee/ScholkopfLBKKGB21, DBLP:conf/iclr/MitrovicMWBB21}, 
we propose a Multi-Task Causal Representation Learning (\framework) framework, aiming to represent the multi-task knowledge via a set of disentangled neural modules instead of a single encoder, and learn the task-to-module causal relationship jointly. We adopt de-correlation and sparsity regularization over popular Mixture-of-Expert (MoE) architecture~\citep{DBLP:conf/iclr/ShazeerMMDLHD17}.
The most critical and challenging step is to learn the causal graph in the MTL setup, which requires distinguishing the genuine causal correlation from spurious ones for all tasks. 
Motivated by the recent studies that invariance could lead to causality~\citep{DBLP:journals/corr/abs-2002-04692, DBLP:journals/corr/abs-2008-01883}, we propose to penalize the variance of gradients w.r.t. causal graph weights across different distributions.
On a high level, this invariance regularization encourages the causal graph to assign higher weights to the modules that are consistently useful. In contrast, the modules encoding spurious knowledge that cannot consistently achieve graph optimality are assigned lower weights and be discarded by task predictors.


We evaluate our method on existing MTL benchmarks, including Multi-MNIST, MovieLens, Taskonomy, CityScape, and NYUv2. 
For each dataset, to mimic distribution shifts, we adopt some attribute information given in the dataset, such as the released time of the movie or district of a building, to split train/valid/test datasets. The results show that \framework could consistently enhance the MTL model’s performance by 5.5$\%$ on average, and outperform both the MTL optimization and robust machine learning baselines. We also conduct case studies to show that \framework indeed alleviate spurious correlation problem in MTL setup.


The key contributions of this paper are as follows:
\begin{compactitem}
\item We are the first to analyze spurious correlation problem in MTL setup, and point out several key challenges unique to MTL with theoretical and empirical analysis.
\item We propose \framework with MTL-specific invariant regularizers to elleviate spurious correlation problem, and enhances the performance on several MTL benchmarks.
\end{compactitem}

\section{Analyzing Spurious Correlation in MTL}\label{problem}
To systematically analyze the spurious correlation problem in MTL, we first assume that data and task labels are generated by ground-truth causal mechanisms described in~\citet{DBLP:conf/icml/SuterMSB19}. 
We denote $X$ as the variable of observed data, and each data is associated with $K$ latent generative factors $\mathbb{F} = \{F_i\}_{i=1}^{K}$ representing different semantics of the data (e.g., color, shape, background of an image). We follow ~\cite{DBLP:journals/pieee/ScholkopfLBKKGB21} to assume that the data $X$ is generated by disentangled causal mechanisms $P(X|F_i)$
, such that $P\big(X | \mathbb{F} \big) = \prod_{i=1}^K P\big(X | F_i \big)$.




As $\mathbb{F}$ represents high-level knowledge of the data, we could naturally define task label variable $Y_t$ for task $t$ as the cause of a subset of generative factors. We denote $\mathbb{F}^C_t$ as a subset of causal feature variables within $\mathbb{F}$ that are causally related to each task variable $Y_t$, and we could define $\mathbb{F}^S_t = \mathbb{F} \setminus \mathbb{F}^C_t$ as a subset of non-causal feature variables to task $t$, such that $P(Y_t | \mathbb{F}) = P(Y_t | \mathbb{F}^C_t)$.
In other words, changing the values of any non-causal factors in $\mathbb{F}^S_t$ does not change the conditional distribution.


Note that the discussion so far is based on the assumption that the ground-truth causal generative model is known. In a real-world learning setting, however, we are only given a supervised dataset $(X,Y)$ without access to generative factors $\mathbb{F}$. To solve the task, a neural encoder $\Phi(\cdot)$ is required to extract representation $\mathbf{Z}$ from the data that encodes the information about the causal factors, on top of which a task predictor $f(\cdot)$ could predict the label.



\begin{figure}
\begin{floatrow}
\ffigbox{%
  \includegraphics[width=0.8\columnwidth]{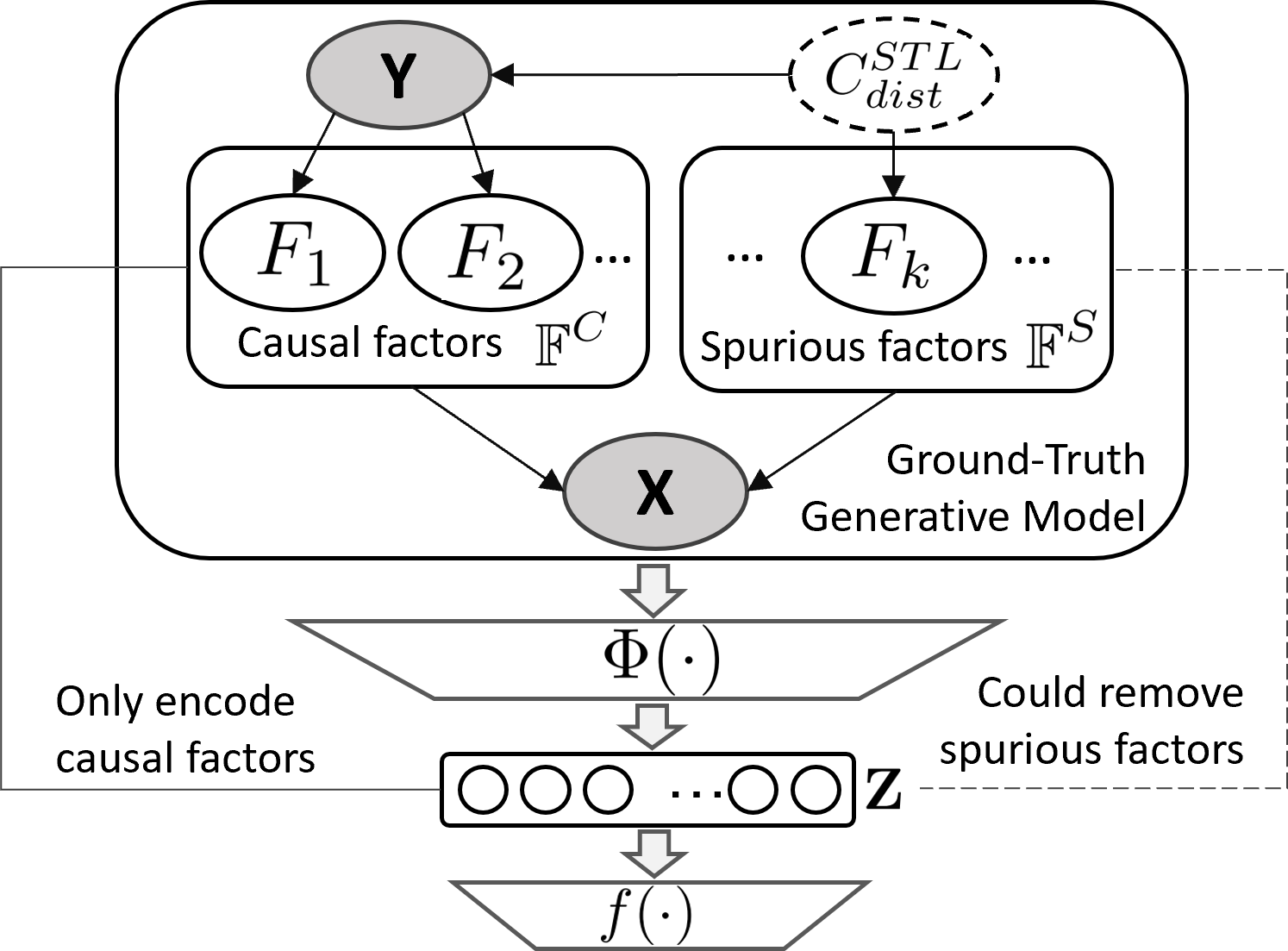}
}{%
  \caption{Spurious correlation in \textbf{Single-Task Learning} is mainly caused by factor-label confounders $C_{dist}^{STL}$. We could remove spurious factors $\mathbb{F}^S$ from representation $Z$.}\label{fig:single}
}
\ffigbox{
    \includegraphics[width=0.9\columnwidth]{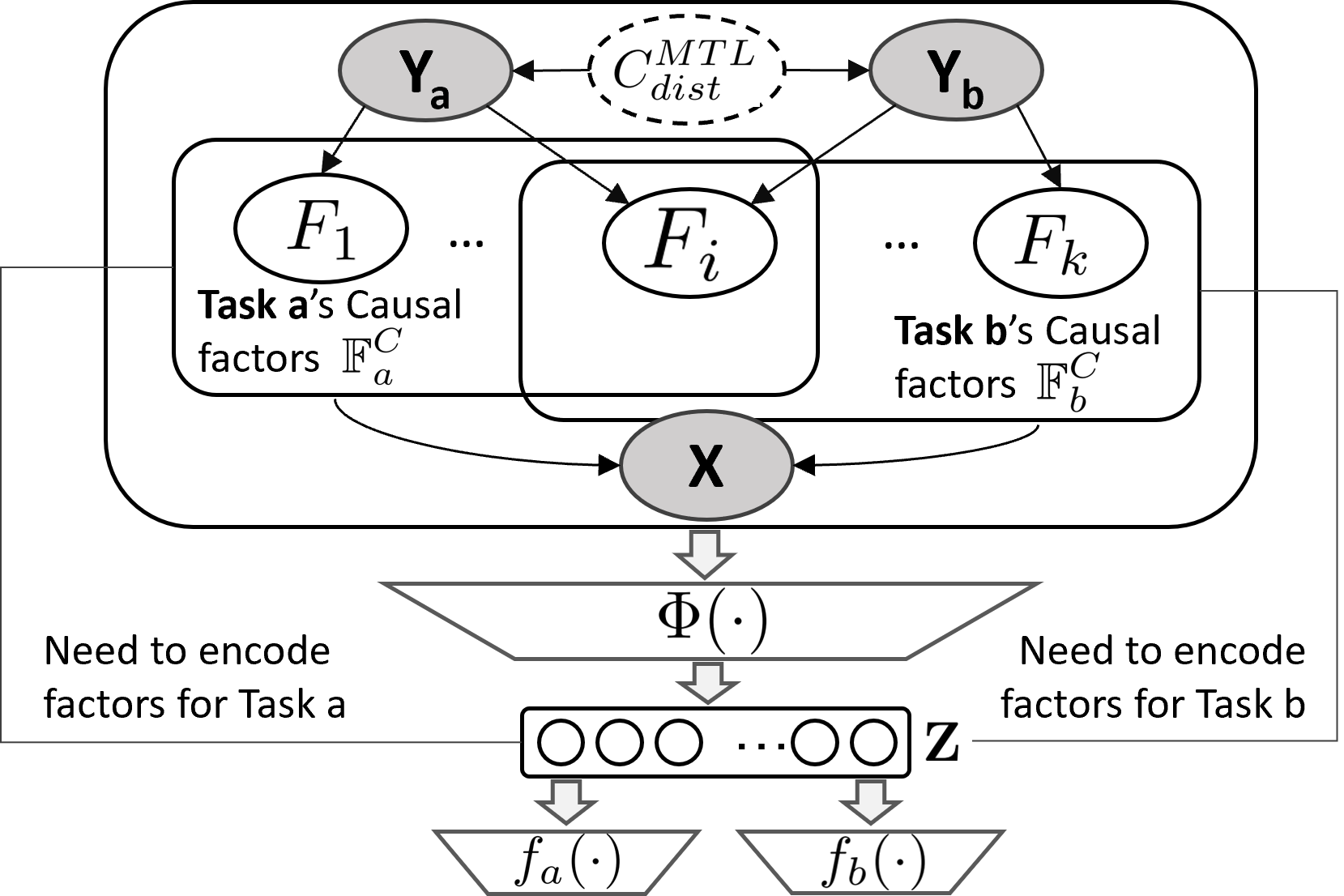}
}{%
   \caption{Spurious correlation in \textbf{Multi-Task Learning} could be caused by label-label confounders $C_{dist}^{MTL}$. Factors for both tasks $\mathbb{F}_a^C$ and $\mathbb{F}_b^C$ need to be encoded and potentially spurious.}\label{fig:multi}
}
\end{floatrow}
\end{figure}

\subsection{Spurious Correlation Problem}

Based on the ground-truth generative model, an ideal predictor for each task should only utilize the causal factors, and keep invariant to any intervention on non-causal factors. 
However, in real-world problems, it is hard to achieve an invariant predictor due to the spurious correlation issue due to unobserved confounders $C_{dist}$~\citep{LopezPaz2016FromDT}.
Formally, confounders are variables that influence the two connected variables’ correlation, and such correlation could change under different distribution (different value of $C_{dist}$), thus the model exploiting such spurious correlation will fail to generalize. Below we summarize the differences of spurious correlation problems for single-task and multi-task learning settings:

\paragraph{Single-Task Learning (STL).}
As illustrated in Figure~\ref{fig:single}, the label-factor confounders for single task learning $C_{dist}^{STL}$ connects non-causal factors $F \in \mathbb{F}^S$ and task label $Y$, bringing in spurious correlation. For example, temperature could confound crime and ice cream consumption. When the weather is hot, both crime rates and ice cream sales increase, but these two phenomena are not causally related. Based on the proof by \citet{DBLP:conf/iclr/NagarajanAN21, DBLP:conf/fat/KhaniL21}, such spurious correlation could lead the model to use non-causal factors, and thus hurt generalization performance.

\paragraph{Multi-Task Learning (MTL).}
In the MTL setting, there exist several unique challenges to handle spurious correlation. First, the risk of having non-causal features is higher. As is illustrated in Figure~\ref{fig:multi}, the shared encoder $\Phi$ needs to encode all the factors causally related to each task in the representation $Z$. Therefore, for each task, all non-overlapping factors from other tasks could be potentially spurious. Second, besides the standard label-factor confounders $C_{dist}^{STL}$ for each single task introduced above, we define label-label confounders $C_{dist}^{MTL}$ connecting multiple tasks' label $\{Y\}$. Such confounder is unique to MTL setting.

As an example, consider two binary classification tasks, with $Y_a$ and $Y_b$ as variables from $\{ \pm 1\}$ for task label. The two labels' correlation $P(Y_a = Y_b) = m_C$ could change with different confounder $C_{dist}^{MTL}=C$. We assume the two tasks have non-overlapping factors $F_a$ and $F_b$ drawn from Gaussian distribution.
We then show MTL model with both two factors as input will utilize non-causal factors:

\begin{propose}\label{pro:spur}
Given $m_C \neq 0.5$, the Bayes Optimal per-task classifier has non-zero weights to non-causal factor. Given $m_C = 0.5$ and limited training dataset, the trained per-task classifier will assign non-zero weights to non-causal factor as noise.
\end{propose}
Detailed proof is in Appendix~\ref{seq:proof}. Therefore, in this linear classification example, when we deploy the trained model to a new distribution with changed label-label confounder $C_{dist}^{MTL}$, the model trained by MTL that utilizes non-causal factors generalize relatively worse. On the contrary, the model trained by STL don't need to encode all causal factors from two tasks. Assuming there is no task-label confounder $C_{dist}^{STL}$ in each task's dataset, the trained model could remove non-causal factors from representation.

\subsection{Empirical Experiments}\label{sec:analysis}

In the following, we conduct experiments to validate the claims. As there is no existing MTL datasets specifically designed to analyze spurious correlation problem, we construct synthetic Multi-SEM~\citep{DBLP:conf/iclr/RosenfeldRR21} and Multi-MNIST~\citep{DBLP:journals/tiis/HarperK16} datasets with known causal structure to study whether the model trained by MTL indeed exploits more non-causal factors, and how the spurious correlation influences multi-task generalization. Dataset details are in Appendix~\ref{sec:syntheticdataset}.


\paragraph{Spurious Score.}
As we know the ground-truth causal structure for the two datasets, we could quantify how much a model utilizes the non-causal factors. Following the gradient saliency map proposed by~\citet{DBLP:journals/corr/SimonyanVZ13}, we calculate the average absolute gradients w.r.t each factor as $Grad(F) = \sum_{(x(\mathbb{F}),y) \in D} \Big\lvert\frac{\partial\big( f(\Phi(x))[y]\big)}{\partial F}\Big\rvert$, which measures how much a model leverage this factor to make prediction. 
We then define the spurious score $\rho_{spur}$ as the proportion of average gradients over non-causal feature $\rho_{spur} = \frac{\sum_{F \in \mathbb{F}^S}  Grad(F)}{\sum_{F \in \mathbb{F}} Grad(F)}$.


\begin{figure}
\begin{floatrow}
\ffigbox{%
  \includegraphics[width=1.0\linewidth]{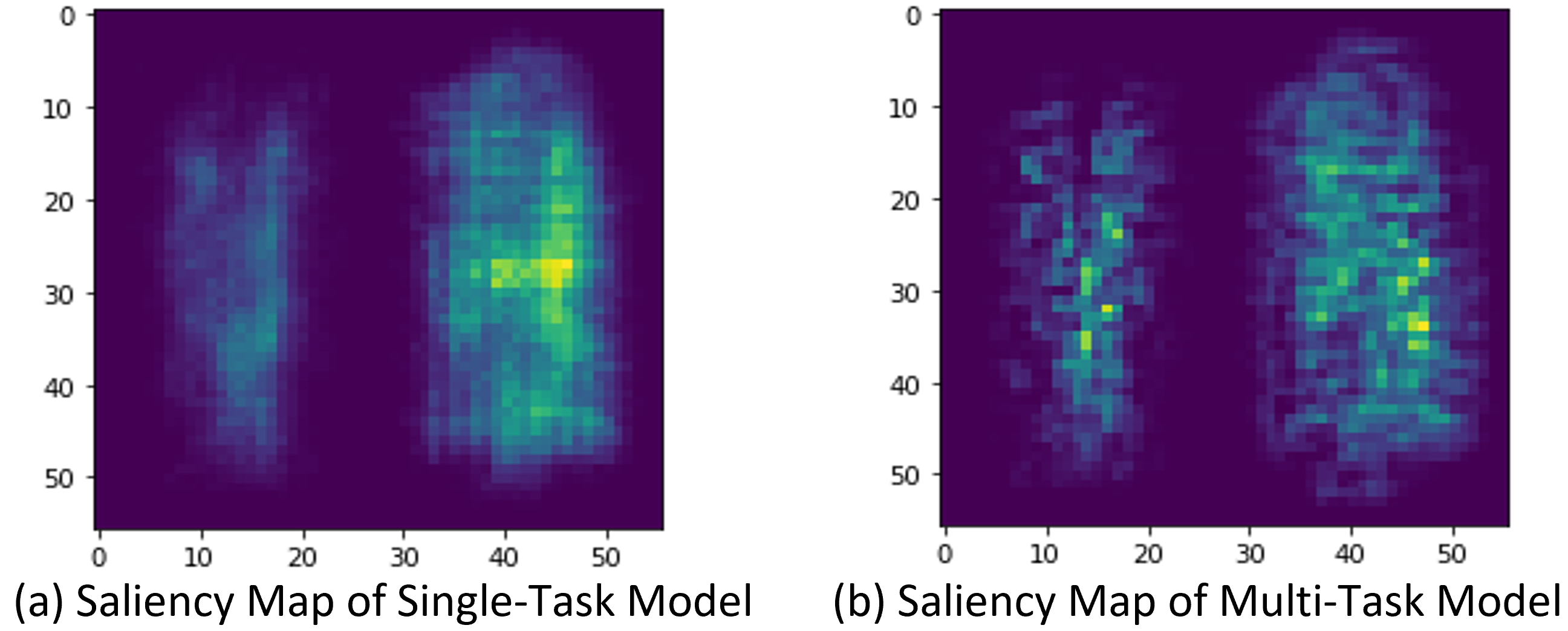}
}{%
  \caption{The gradient saliency map of right-side digit classifier. The model trained by MTL exploits left pixels (spurious) more.}
    \label{fig:vis}
}
\capbtabbox{
    \begin{tabular}{l|cc|cc} 
    \toprule
    & \multicolumn{2}{c|}{\textbf{Multi-SEM}} & \multicolumn{2}{c}{\textbf{Multi-MNIST}} \\
    & STL & MTL & STL & MTL\\ \midrule
    Acc$_{train}$ & 0.931 & 0.936 & 0.981 &  0.987 \\
    Acc$_{val}$ & 0.906 & 0.882 & 0.874 &  0.846 \\
    $\rho_{spur}$ & 0.128 & 0.246 & 0.261 &  0.328 \\
    \bottomrule
    \end{tabular}
}{%
  \caption{Empirical results of multi-task (MTL) and single-task learning (STL) model on synthetic datasets with changing $C_{dist}^{MTL}$.}
    \label{tab:emp}
}
\end{floatrow}
\end{figure}


\paragraph{Empirical Results.}
We train a shared-bottom model via Multi-task learning (MTL) and single-task learning (STL) over the two datasets and report both the training and test accuracy with spurious ratio $\rho_{spur}$ in Table~\ref{tab:emp}. As illustrated, the test accuracies of MTL for both Multi-SEM and Multi-MNIST datasets are both worse than STL. The training accuracies of MTL are very similar to STL, meaning that the performance drop is not due to the optimization difficulty that many previous works try to address. The spurious ratio $\rho_{spur}$ of MTL is much higher than the STL, which means that it exploits more non-causal factors. To give a more straightforward illustration, we plot the gradient saliency map of the right-side digit classifier for Multi-MNIST in Figure~\ref{fig:vis}. The model trained by MTL utilizes more left-side pixels, which are non-causal to the final prediction. We also show the results of Multi-SEM with more than 2 tasks in Appendix~\ref{sec:more}. These findings support our hypothesis that with spurious correlation caused by label-label confounder $C_{dist}^{MTL}$, models trained by MTL is more prone to leverage non-causal knowledge than STL, and thus influence generalization performance.

\section{Method}


Based on the previous analysis of the spurious correlation problem in MTL, we now introduce a Multi-Task Causal Representation Learning (\framework) framework with the goal that the per-task predictor only leverages the causal knowledge instead of spurious correlation. The high-level idea of the framework is to reconstruct the ground-truth causal mechanisms introduced in section~\ref{problem} through end-to-end representation learning. 
To accomplish this goal, the framework aims to 1) model multi-task knowledge via a set of disentangled neural modules; 2) learn the task-to-module causal graph that is optimal across different distributions.
With the correct causal graph as routing layer, per-task predictor only utilizes outputs from causally-related modules, thus alleviating the spurious correlation problem.
We introduce the two crucial designs as follows.

\subsection{Modelling via Disentangled Neural Modules}
In order to alleviate spurious correlation, an ideal MTL model should learn the multi-task knowledge in the shared representation while identifying which part of the knowledge is causally related to each task. However, directly conducting causal discovery is impossible if all the knowledge is fused in a single shared encoder. Thus, we seek to adopt a modularized architecture in which each module encodes disentangled knowledge, and thus enable modeling causal relationship between task and modules. 
We adopt Multi-gate Mixture-of-Experts (MMoE)~\citep{DBLP:conf/kdd/MaZYCHC18}, a variant of MoE~\citep{DBLP:conf/iclr/ShazeerMMDLHD17} architecture tailored for MTL setting, as our underlying model.
Specifically, we have $K$ different neural modules as shared encoders $\Phi = \big[\Phi_i(\cdot)\big]_{i=1}^K$. 
Given a batch of input data $\mathbf{X} = \{x_n\}_{n=1}^B$ with batch size $B$, we extract $k$ representations via different neural modules, i.e., $\mathbf{Z}_i = \Phi_i(X) \in \mathbb{R}^{B \times d}$.
Based on sparsity assumption of the causal mechanisms~\cite{DBLP:conf/icml/ParascandoloKRS18,DBLP:conf/iclr/BengioDRKLBGP20, lachapelle2021disentanglement}, only a few modules should be causally related to each task. Therefore, on top of the learned neural modules, we learn a task-to-module routing graph, aiming to estimate which module is causally related to each task. We model the bipartite adjacency (a.k.a. bi-adjacency)
matrix $A=\text{sigmoid}(\theta) \in [0,1]^{T \times K}$ by applying sigmoid over a learnable parameter $\theta$ to enforce the range constraint. Note that original MMoE adopts softmax to get gate vector, which encourages only a small portion of modules being utilized for each task. Our graph modelling allows multiple modules utilized for each task. 
With the correct graph weights $A$ as routing layer, 
we could utilize only the causally related modules and make predictions with per-task predictor $f_t(\cdot)$ as $\hat{Y}_t(\mathbf{X}) = f_t\big(\sum_{i}A_{t,i} \cdot \Phi_i(\mathbf{X})\big)$.

\paragraph{Disentangling Modules.}
One of the main properties of the causal mechanisms we introduced in section~\ref{problem} is disentanglement, such that each factor represents a different view of the data, and changing the value of one factor does not influence the others. If without explicit constraints, the learned modules' outputs could still be correlated and hinder the causal structure learning. Therefore, we need to add regularization to disentangle these modules during training.


Most existing disentangled representation learning methods are under the generative modeling framework, e.g. VAE~\citep{DBLP:conf/iclr/HigginsMPBGBML17} or GAN~\citep{DBLP:conf/nips/ChenCDHSSA16}. However, \citet{DBLP:conf/icml/LocatelloBLRGSB19} argues that without explicit supervision, it is hard for generative models to learn correct disentangled factors.
We therefore only borrow the regularization methods utilized in existing generative disentangled representation works~\citep{DBLP:journals/corr/CheungLBO14, DBLP:journals/corr/CogswellAGZB15} to directly penalize the correlation of learned modules. 
Specifically, we regularize the in-batch Pearson correlation $\rho(\mathbf{Z}_i, \mathbf{Z}_j)$ between every pair of output dimensions from different representation matrices $\mathbf{Z}_i$ and $\mathbf{Z}_j$, as:
\begin{align}
\rho(\mathbf{Z}_i, \mathbf{Z}_j) = \frac{Cov(\mathbf{Z}_i, \mathbf{Z}_j)}{\sqrt{Cov(\mathbf{Z}_i, \mathbf{Z}_i)} \sqrt{Cov(\mathbf{Z}_j, \mathbf{Z}_j)}},\  \text{where} \  Cov(\mathbf{Z}_i, \mathbf{Z}_j) = \big[\mathbf{Z}_i - \overline{\mathbf{Z}_i}\ \big]^T\big[\mathbf{Z}_j - \overline{\mathbf{Z}_j}\ \big].
\end{align}    
By minimizing the Frobenius norm of the correlation matrix $\rho$ for every two different representation pairs, we could enforce the encoder $\Phi$ to extract disentangled representations.
\begin{align}
    \mathcal{L}_{decor}(\Phi) = \lambda_{decor} \cdot {\sum_{i = 1}^{k}\sum_{j = i+1}^k}\Big\lVert \rho\big(\Phi_i(X), \Phi_j(X)\big)  \Big\rVert_F^2 \label{eq:decor}.
\end{align}
\paragraph{Task-to-Module Graph Regularization.}
Based on sparsity assumption of the causal mechanisms~\citep{DBLP:conf/icml/ParascandoloKRS18,DBLP:conf/iclr/BengioDRKLBGP20, lachapelle2021disentanglement}, each task is causally related to only a few modules.
To learn the graph structure, existing works~\citep{DBLP:conf/nips/ZhengARX18, DBLP:journals/corr/abs-1910-08527, DBLP:conf/iclr/LachapelleBDL20} propose to to fit structural equation model (SEM) with sparsity regularization over the graph weights. We adopt a similar sparse regularization with an entropy balancing term~\citep{hainmueller2012entropy} over the bi-adjacency matrix $A$ weights of the task-to-module routing graph:
\begin{align}
     \mathcal{L}_{graph}(A) = \lambda_{sps} \cdot ||A||_1 - \lambda_{bal} \cdot \text{Entropy}\Big(\frac{\sum_t A_{t, *}}{\sum_{t, i} A_{t, i}}\Big) \label{eq:graph}.
\end{align}     
Note that the entropy term aims at keeping the causal weights for each module $i$ summing over all the tasks to be balanced. This could help avoid degenerate solutions in which only a few modules are utilized. By combining the two regularizations in Eq.(\ref{eq:decor}) and Eq.(\ref{eq:graph}) with per-task supervised risk term $R_t\big( \Phi, A_t, f_t \big) = \sum_{(\mathbf{X}, \mathbf{Y}_t) \in \mathcal{D}} L_t\big(\hat{Y}_t(\mathbf{X}), \mathbf{Y}_t\big)$, we get the regularized loss as:
\begin{align}
    &\mathcal{\tilde{L}}(\Phi, A, f) = \sum_{t \in \mathcal{T}} R_t\big(\Phi, A_t, f_t \big) + \mathcal{L}_{decor}(\Phi) + \mathcal{L}_{graph}(A).
\end{align}

\subsection{Causal Learning via Graph-Invariant Regularization}

It is critical and challenging to learn the correct causal graph, which requires distinguishing the true causal correlation from spurious ones. Motivated by the recent studies of robust machine learning that a predictor invariant to multiple distributions could learn causal correlation~\citep{DBLP:journals/corr/abs-2002-04692, DBLP:journals/corr/abs-2008-01883}, we assume the true causal relationship to be optimal across different distributions. To do so, we assume to have access to multiple slices of datasets collected from different environments $e \in \mathcal{E}$ in which the confounder $C_{dist}^{MTL}$ that controls task correlation might change. For example, one natural choice is to consider train/valid dataset split (the setting we utilize in experiment), or assume the training set is split into multiple slices with different attributes. We desire the task-to-module graph weights $A$ and per-task predictor $f_t$ to be optimal across all environments $e \in \mathcal{E}$. Formally, we aim to solve the following bi-level optimization problem:  
\begin{align}
    \min_{\Phi, A, f}\  \mathcal{\tilde{L}}(\Phi, A, f) \  \ \ \text{s.t.}  \  \  A_t, f_t \in \arg \min_{A, f} R^e_t\big(\Phi, A, f \big), \forall \  t \in \mathcal{T}, e \in \mathcal{E}  \label{eq:opt}.
\end{align} 
where $R^e_t$ denotes the risk over data slice in environment $e$. This optimization problem could be regarded as a multi-task version of IRM. Based on Theorem 9 described in~\citet{DBLP:journals/corr/abs-2002-04692}, by enforcing invariance over a sufficient number of environments that exhibit distribution shifts (i.e., changes of confounder $C_{dist}^{MTL}$), per-task predictors should only utilize modules that are consistently helpful to the task, and assign zero weights to modules that encode non-causal factors to the task, and thus alleviate spurious correlation and help out-of-distribution generalization. Even if all data are sampled from the same distribution and there are no distribution shifts, invariance could also help eliminate noisy correlation due to the limited training dataset and help in-distribution generalization.

\paragraph{Invariant Optimality of Task-to-Module Graph for MTL.} As discussed in IRM, the bi-leveled optimization problem in Eq.(\ref{eq:opt}) is highly intractable, especially with complex and non-linear $\Phi$. To implement a practical optimization objective, IRM proposes to softly regularize the gradient of the task-predictor at different environments to enforce it to be optimal:
\begin{align}
    \min_{\Phi, A, f}\ \Big( \mathcal{\tilde{L}}(\Phi, A, f) + \sum_{t \in \mathcal{T}}\sum_{e \in \mathcal{E}}\Big\lVert \nabla_{A=A_t, f=f_t} R^e_t\big(\Phi, A, f \big) \Big\rVert^2 \Big) \label{eq:IRM}.
\end{align} 

However, as is discussed in IRM paper, if the complexity of a task-predictor $f$ is much larger than the number of environments, it could learn an over-fitted solution that makes gradient zero but does not achieve invariance. IRM adopts a fixed all-one vector as predictor to reduce complexity. This approach \textbf{is not applicable to MTL setup}, as the optimal task-predictors $f_t^*$ for different task $t$ could be very distinctive and complex, and we cannot use a fixed uniform predictor for all tasks.

To strike a balance between invariance and complexity of multi-task predictors, we propose only to regularize the gradient of the task-to-module routing graph while assuming the complex predictor $f_t$ for each task is fixed at each iteration. We call this modification as \textbf{Graph-Invariant Risk Minimization (G-IRM)}, which is designed specifically to MTL setup:
\begin{align}
&\min_{\Phi, A, f}\  \Big( \mathcal{\tilde{L}}(\Phi, A, f) + \lambda_{G\text{-}IRM} \cdot  \mathcal{L}_{G\text{-}IRM}(\Phi,  A | f) \Big) \label{eq:problem}.
\end{align} 
By adopting the similar gradient penalty term as adopted in IRM, we define $\mathcal{L}_{G\text{-}IRM}^{Norm}$ as:
\begin{align}
\mathcal{L}_{G\text{-}IRM}^{Norm}(\Phi,  A | f) = \sum_{t \in \mathcal{T}}\sum_{e \in \mathcal{E}}\Big\lVert \nabla_{A = A_t} R^e_t\big(\Phi, A, f_t \big) \Big\rVert^2.
\end{align} 
As we assume $f_t$ is fixed for invariance regularization term $\mathcal{L}_{G\text{-}IRM}^{Norm}$, we only calculate gradient and optimize for $\Phi$ and $A$, but not updating $f_t$. This could avoid the over-parametrized predictor $f_t$ finding a trivial solution to achieve zero gradients instead of learning the correct causal correlation. Similar trick is utilized in~\citep{DBLP:conf/iclr/AhmedBSC21}.
Note that the gradient w.r.t each graph weight means whether a module could help reduce the risk for this task. Therefore, by penalizing the invariance regularization, the modules containing non-causal factors will be assigned zero weights.

In the experiments, we observe that at the early optimization stage, the model has non-zero gradients for all parameters, including the graph weights, thus directly regularizing the gradient norm might influence the optimization. Therefore, we propose a modified version of gradient regularization $\mathcal{L}_{G\text{-}IRM}^{Var}$ that penalizes the variance of the task-to-module graph's gradient on different environments:
\begin{align}
    &\mathcal{L}_{G\text{-}IRM}^{Var}(\Phi,  A | f) = \sum_{t \in \mathcal{T}} \sum_{e \in \mathcal{E}} \frac{1}{|\mathcal{E}|} \Big\lVert \nabla_{A = A_t} R^e_t\big(\Phi, A, f_t \big) - \text{Avg}_e\Big(\nabla_{A = A_t} R^e_t\Big) \Big\rVert^2.
\end{align} 
By minimizing $\mathcal{L}_{G\text{-}IRM}^{Var}$, we force all the learned modules to have similar gradients across different environments, and not overfit only to some of the environments. It still allows some modules to have non-zero gradients as long as it's the same across environments, and relies on loss term $\mathcal{\tilde{L}}$ to update these weights, while $\mathcal{L}_{G\text{-}IRM}^{Norm}$ forces all gradient to be zero. Therefore, $\mathcal{L}_{G\text{-}IRM}^{Var}$ is a loose regularization that not influences the overall optimization too much. It shares similar intuition of REx~\citep{DBLP:conf/icml/KruegerCJ0BZPC21} that penalizes risk variance, while $\mathcal{L}_{G\text{-}IRM}^{Var}$ penalize gradient variance. We provide pseudo-code of \framework framework in Appendix~\ref{sec:code}.

\section{Experiment}

In this section, we evaluate whether \framework could benefit the performance of MTL models on existing benchmark datasets, and study whether it could indeed alleviate spurious correlation.


\paragraph{Experimental Setup.} 
One key ingredient of our \framework is to achieve the optimality of causal graph over different distributions. However, we might not access multiple environmental labels in most real-world multi-task learning datasets. Therefore, we adopt a more realistic setup, such that we only assume to have a single validation set that contains unknown distribution shifts (i.e. change of confounder $C_{dist}^{MTL}$) compared to the training dataset. We thus could utilize training and valid sets as two environments to calculate invariance regularization, while we only utilize the training set to calculate task loss to avoid the task predictor overfits. Note that in this way, our method could get access to the label information in the validation set. To avoid the possibility that the performance improvement is brought by additional label, for all the other baseline methods, we also add the validation data into the training set to calculate task loss and learn MTL model.

\paragraph{Dataset.}
We choose five widely-used real-world MTL benchmark datasets, 
i.e., Multi-MNIST~\citep{mulitdigitmnist}, MovieLens~\citep{DBLP:journals/tiis/HarperK16}, Tasknomy~\citep{DBLP:conf/cvpr/ZamirSSGMS18}, NYUv2~\citep{DBLP:conf/eccv/SilbermanHKF12} and CityScape~\citep{DBLP:conf/cvpr/CordtsORREBFRS16},
and try to determine train/valid/test split such that there exist distribution shifts between these sets. Dataset details are in Appendix~\ref{sec:dataset}. Note that except NYUv2, our data split is the same as the default split settings of these datasets, which also try to test model's capacity to generalize across domains.






\paragraph{Baselines.}
As \framework is a regularization framework built upon modular MTL architecture (in this paper we choose MMoE as instantiation, but it can be applied to other modular networks),
we mainly compare with two gradient-based multi-task optimization baselines: \textbf{PCGrad}~\citep{DBLP:conf/nips/YuK0LHF20} and \textbf{GradVac}~\citep{DBLP:conf/iclr/WangTF021}.
We also compare with two domain generalization baselines: \textbf{IRM}~\citep{DBLP:journals/corr/abs-2002-04692} and \textbf{DANN}~\citep{DBLP:journals/jmlr/GaninUAGLLML16}. For IRM we adopt different per-task predictors instead of all-one vector to adapt MTL setup, and calculate penalty via Eq. (\ref{eq:IRM}).

\paragraph{Hyper-Parameter Selection.}
For a fair comparison, all methods are based on the same MMoE architecture. 
Our methods contain a lot of hyper-parameters, including some model specific ones such as number of modules ($K$) and regularization specific ones. To avoid the case that performance improvement is caused by extensive hyper-parameter tuning, we mainly search optimal model hyper-parameter on Vanilla MTL setting, and use for all baselines. For regularization specific parameters, we take Multi-MNIST, the simplest dataset among the testbed, to find a optimal combination, and use for all other datasets. Detailed selection procedure and results are shown in Appendix~\ref{sec:hyper}.


\subsection{Experiment Results}

\begin{table*}[t!]
\footnotesize
\centering
\begin{tabular}{l|cccccc} \toprule
 \textbf{Methods}    & \textbf{Multi-MNIST} & \textbf{MovieLens}  & \textbf{Taskonomy}  & \textbf{CityScape}  & \textbf{NYUv2} & \textbf{Avg.}\\ \midrule
 Vanilla MTL & \multicolumn{6}{c}{(---baseline to calculate relative improvement---)}\\ 
 Single-Task Learning & +3.3\% & +0.2\% & -2.5\%& -2.4\% & -12.2\%  & -2.7\%\\ \midrule
 MTL + PCGrad & +4.5\% & +0.2\% & +3.1\%& +2.1\% & +7.4\%  & +3.5\% \\
 MTL +  GradVac & +4.6\% & +0.3\% & +3.5\%& +2.1\% & +7.2\% & +3.5\% \\ \midrule 
 MTL + DANN & +4.1\% & +0.4\% & +1.2\% & +0.3\% & -0.4\% & +1.1\%\\
 MTL + IRM & +5.0\% & +0.4\% & +1.1\%& +0.6\% & -0.1\% & +1.4\%\\ \midrule
MT-CRL w/o $\mathcal{L}_{G\text{-}IRM}$ & +5.9\% & +0.2\% & +3.2\% & +1.5\% & +4.3\%& +3.0\% \\
MT-CRL with $\mathcal{L}_{G\text{-}IRM}^{Norm}$ & +7.8\% & +1.0\% & +6.5\% & \textbf{+2.9\%} & +8.0\%& +5.2\%  \\
MT-CRL with $\mathcal{L}_{G\text{-}IRM}^{Var}$ & \textbf{+8.1\%} & \textbf{+1.1\%} & \textbf{+7.1\%} & +2.8\% & \textbf{+8.2\%} & \textbf{+5.5\%} \\ \bottomrule
\end{tabular}
\caption{Relative Performance improvement of different multi-task learning (MTL) strategies compared to vanilla MTL baseline. Detailed results for each task are shown in Table~\ref{tab:res1}-\ref{tab:res5} in Appendix~\ref{result}.}
\label{tab:e2e}
\end{table*}

As each task has a different evaluation metric and cannot be directly compared, we calculate the relative performance improvement of each method compared to vanilla MTL, and then average the relative improvement for all tasks of each dataset.  As summarized in Table~\ref{tab:e2e}, the average  improvement of \framework with $\mathcal{L}_{G\text{-}IRM}^{Var}$ is $5.5\%$, significantly higher than all other baseline methods.
The most critical step of \framework is to learn correct causal graph. We therefore report \framework with different invariance regularization. As is shown in the last block, $\mathcal{L}_{G\text{-}IRM}^{Var}$ achieve better results for most datasets than $\mathcal{L}_{G\text{-}IRM}^{Norm}$, while removing the invariance regularization could significantly drop the relative performance. Compared to IRM which calculate gradient and update per-task predictors, \framework uses disentangled modules and G-IRM to avoid overfitting to achieve invariance. Results show that for datasets with large amount of tasks, e.g., Taskonomy and NYUv2, \framework significantly outperform IRM, showing the modification is more suitable for MTL setup.

\begin{figure}
\begin{floatrow}
\footnotesize
\capbtabbox{
    \begin{tabular}{cc|cc|cc} \toprule
    \multicolumn{2}{c|}{\textbf{Disentangled Reg.}} & \multicolumn{2}{c|}{\textbf{Graph Reg.}} & Multi-MNIST  \\
    $\mathcal{L}_{decor}$ & $\mathcal{L}_{\beta\text{-VAE}}$   &  $\mathcal{L}_{sps}$    & $\mathcal{L}_{bal}$   & Accuracy \\ \midrule
    \cmark&\xmark  &\cmark &\cmark   & 0.915 $\pm$ 0.018  \\
    \xmark&\cmark  &\cmark &\cmark   & 0.896 $\pm$ 0.024   \\
    \xmark&\xmark  &\cmark &\cmark   & 0.882 $\pm$ 0.020   \\ \midrule
    \cmark&\xmark  &\xmark &\xmark  & 0.891 $\pm$ 0.016   \\
    \cmark&\xmark  &\xmark &\cmark   & 0.903 $\pm$ 0.017   \\
    \cmark&\xmark  &\cmark &\xmark   & 0.908 $\pm$ 0.021   \\ 
    \bottomrule
    \end{tabular}
}{%
 \caption{\textbf{Ablation Studies} of disentangled and Graph regularization components in \framework, evaluated on Multi-MNIST dataset. }\label{tab:ablation}
}

\ffigbox{
  \includegraphics[width=1.1\linewidth]{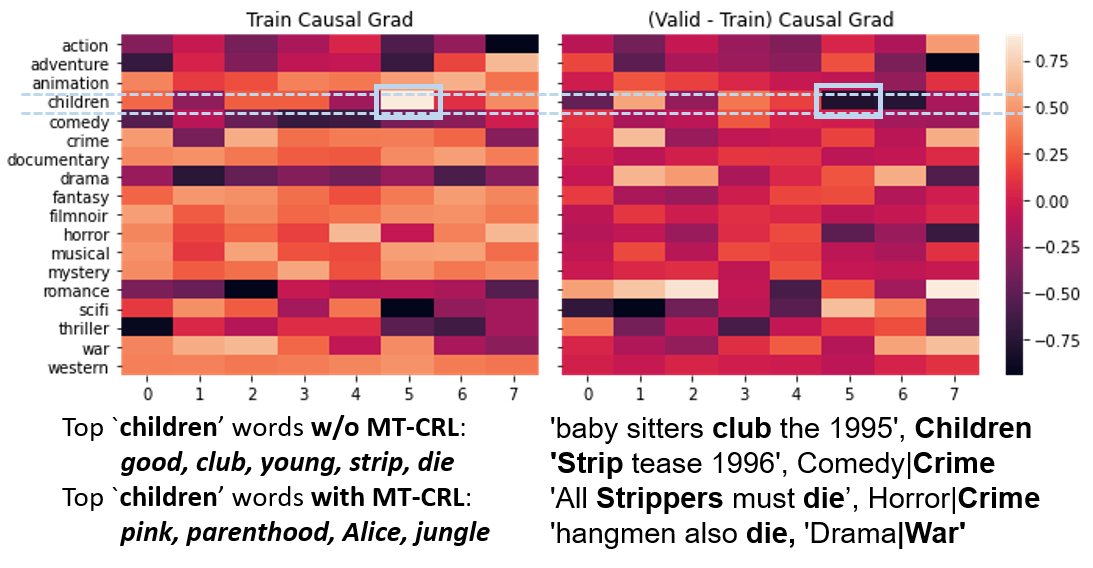}
}{%
  \caption{Task-to-Module gradients of model without MT-CRL show Module 5 is spurious. MT-CRL could help alleviate spurious correlation.} \label{fig:movie}
}

\end{floatrow}
\end{figure}

\begin{figure}[t]
\vspace*{-.2in}
    \centering
    \includegraphics[width=0.9\textwidth]{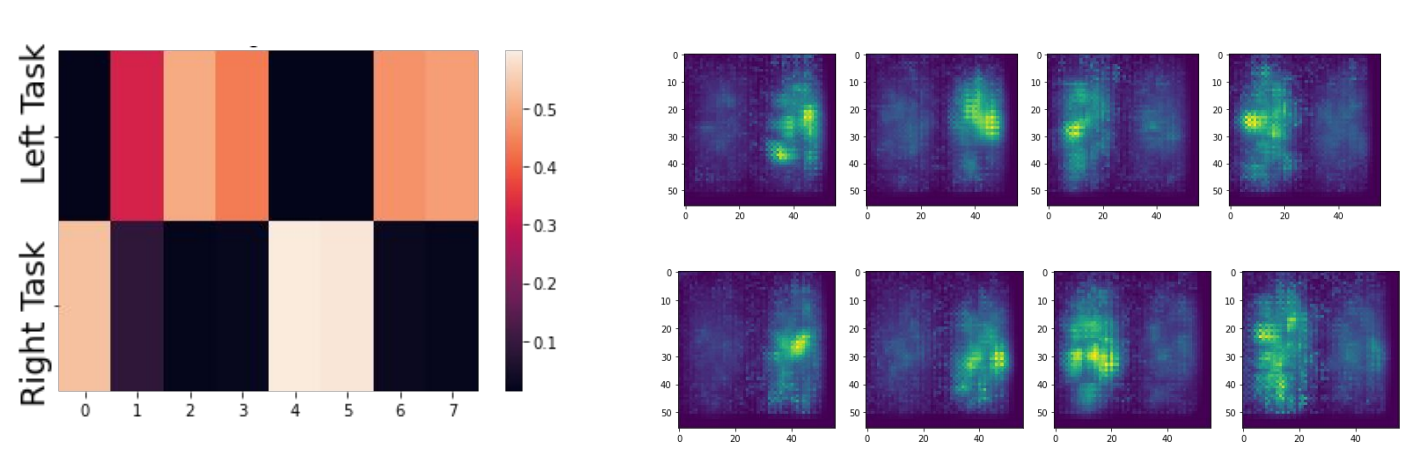}
    \caption{(valid-train) Task-to-Module gradients of model \textbf{with} MT-CRL on Multi-MNIST.}
    \label{fig:mnist_after}
\vspace*{-.15in}
\end{figure}

\paragraph{Ablation Studies.}
We then study the effectiveness of the other two components in \framework, i.e., disentangled and graph regularization. We mainly report the ablation studies on Multi-MNIST in table \ref{tab:ablation} as it's relatively small so that we could quickly get the results of all combinations.

For disentangled regularization, after removing $\mathcal{L}_{decor}$, the performance drops from 0.915 to 0.882, which fits our discussion that we cannot conduct causal learning over entangled modules.
We also explore one classical generative disentangled representation method, i.e., $\beta$-VAE. As shown in the table, the results of using $\beta$-VAE are 0.896, lower than our utilized decorrelation regularization.. We hypothesize that this is probably because not all generative factors are useful for downstream tasks. Generative objectives might compete for the model capacity and in addition, the unused factors could be potentially spurious.

Another key component is graph regularization. After removing both $\mathcal{L}_{sps}$ and $\mathcal{L}_{bal}$, the performance drops to $0.891$. This show that even if invariance regularization could penalize non-causal modules, it would be better to force their weights to be zero via sparsity regularization, and to be non-degenerate via balance regularization. We also conduct ablation studies to remove either $\mathcal{L}_{sps}$ or $\mathcal{L}_{bal}$, and results show both are important, and combining the two could help to achieve the best results.

\paragraph{Case Study.} To show that real-world MTL problem indeed have spurious correlation problem and our \framework could alleciate it, we take MovieLens as an example to conduct case study. Each task is for different movie types, and bag-of-word of movie title is one of the features. 
We calculate the task-to-module gradients $\frac{\partial( f(\Phi(x))[y])}{\partial F}$ 
of the vanilla MMoE model without \framework. We then visualize `train' gradients, which shows how much each module is utilized to fit the training set, and `valid-train' gradients, which shows how generalizable each module is. We find that module 5 is utilized for \textbf{children} movie, but harmful in valid set, indicating it is a spurious feature. We then use Grad-CAM to show that top words of module 5 include \textit{strip} and \textit{die}, which is not relevant to \textbf{children} movies. One possible reason is that some children movies contain the words \textit{club}, which is often co-occurred with \textit{strip} and \textit{die} in \textbf{crime} and \textbf{war} movies. After adding our \framework, the module assigned to `children' movie attends \textit{Pink}, \textit{Parenthood}, \textit{Alice} and \textit{Jungle}.

We then show the (valid-train) Task-to-Module gradients over Multi-MNIST datasets. With \framework, in Figure~\ref{fig:mnist_after}, each module's saliency map only focus on one side of pixels. By looking at each task output's saliency map, which help model to focus only on causal part, compared with Figure~\ref{fig:vis}(b) that have high weights on both. We also show the detailed gradient saliency map and induced task similarity graph of MovieLens, Taskonomy in Appendix~\ref{sec:case}. All these case studies show \framework could indeed alleviate spurious correlation in real MTL problems.

\section{Related Work}

\paragraph{Multi-Task Generalization.}
A deep neural model often requires a large number of training samples to generalize well~\citep{DBLP:conf/icml/AroraDHLW19, DBLP:conf/nips/CaoG19a}. To alleviate the sample sparsity problem, MTL could leverage more labeled data from multiple tasks~\citep{zhang2018overview}.
Most works studying multi-task generalization are based on a core assumption that the tasks are correlated. Earlier research directly define the task relatedness with statistical assumption~\citep{baxter2000model,DBLP:journals/ml/Ben-DavidB08,DBLP:conf/iclr/LampinenG19}. With the increasing focus on deep learning models, recent research decompose ground-truth MTL models into a shared representation and different task-specific layers from a hypothesis family~\citep{DBLP:journals/jmlr/MaurerPR16}. With such decomposition, \citet{DBLP:conf/nips/TripuraneniJJ20} and \citet{DBLP:conf/iclr/DuHKLL21} prove that a diverse set of tasks could help learn more generalizable representation. \citet{DBLP:conf/iclr/0002ZR20} study how covariate shifts influence MTL generalization.
Despite these findings, the core assumption of task relatedness might not be satisfied in many real-world applications~\citep{DBLP:journals/corr/ParisottoBS15, DBLP:journals/corr/abs-2009-00909}, in which tasks could even conflict with each other to compete model capacity, and the generalization performance of MTL could be worse than single-task training.

To solve the task conflict problem, a number of MTL model architectures have utilized modular~\citep{DBLP:conf/cvpr/MisraSGH16, DBLP:conf/cvpr/LuKZCJF17, DBLP:conf/iclr/RosenbaumKR18,DBLP:conf/kdd/MaZYCHC18,DBLP:conf/icml/GuoLU20} or attention-based~\citep{DBLP:conf/cvpr/LiuJD19, DBLP:conf/cvpr/ManinisRK19} design to enlarge model capacity while preserving information sharing. 
Our work is model-agnostic and could be applied to existing architectures to further solve the spurious feature problem. 
Another line of research alleviate task conflict during optimization. Some propose to balance the task weight via uncertainty estimation~\citep{DBLP:conf/cvpr/KendallGC18}, gradient norm~\citep{DBLP:conf/icml/ChenBLR18}, convergence rate~\citep{DBLP:conf/cvpr/LiuJD19}, or pareto optimality~\citep{DBLP:conf/nips/SenerK18}. Others directly modulate task gradients via dropping part of the conflict gradient~\citep{DBLP:conf/nips/ChenNHLKCA20} or project task’s gradient onto other tasks' gradient surface~\citep{DBLP:conf/nips/YuK0LHF20, DBLP:conf/iclr/WangTF021}. 
Though these works successfully facilitate MTL model to converge easier, our analysis show that with spurious correlation, the MTL model with low training loss could still generalize bad. 
Therefore, our proposed \framework that alleviates spurious correlation is orthogonal to these prior works, and could be combined to further improve overall performance.

\paragraph{Spurious Correlation Problem.} 
Due to the selection bias~\citep{DBLP:conf/cvpr/TorralbaE11, DBLP:conf/naacl/GururanganSLSBS18} or unobserved confounding factors~\citep{LopezPaz2016FromDT}, training datasets always contain spurious correlations between non-causal features and task labels, with which trained models often leverage non-causal knowledge and may fail to generalize Out-Of-Distribution (OOD) when such correlation changes~\citep{DBLP:conf/iclr/NagarajanAN21}. 
To solve the spurious correlation problem, some fairness research pre-define a set of non-causal features (e.g., gender and underrepresented identity) and then explicitly remove them from the learned representation~\citep{DBLP:conf/icml/ZemelWSPD13,DBLP:journals/jmlr/GaninUAGLLML16,DBLP:conf/iccv/WangZYCO19}. Another line of robust machine learning research does not assume to know spurious features, but regularize the model to perform equally well under different distribution. Distributionally Robust Optimization (DRO) optimizes worst-case risk~\citep{DBLP:conf/iclr/SagawaKHL20}. Invariant
Causal Prediction (ICP) learns causal relations via invariance testing~\citep{peters2016causal}. Invariant Risk Minimization (IRM) forces the final predictor to be optimal across different domains~\citep{DBLP:journals/corr/abs-1907-02893}. Risk Extrapolation (REx) directly penalizes the variance of training risk in different domains~\citep{DBLP:conf/icml/KruegerCJ0BZPC21}. 
Another line of work aim at learning causal representation~\citep{DBLP:journals/pieee/ScholkopfLBKKGB21}, i.e., high-level variables representing different aspect of knowledge from raw data input. Most of these works try to recover disentangled causal generative mechanisms~\citep{DBLP:conf/icml/ParascandoloKRS18, DBLP:conf/iclr/BengioDRKLBGP20, DBLP:journals/corr/abs-2011-01681, DBLP:conf/iclr/MitrovicMWBB21}. 
Despite the extensive study of spurious correlation in single-task setting, few work discuss it for MTL models. This paper is the first to point out the unique challenges of spurious correlation in MTL setup.

\section{Conclusion}
In this paper, we study spurious correlation problem in the Multi-Task Learning (MTL) setting. 
We theoretically and experimentally shows that task correlation can introduce special type of spurious correlation in MTL, and the model trained by MTL is more prone to leverage non-causal knowledge from other tasks than single-task learning. 
To solve the problem, we propose Multi-Task Causal Representation Learning (\framework) which consists of: 1) a decorrelation regularizer to learn disentangled modules; 2) a graph regularizer to learn sparse and non-degenerate task-to-module graph; 3) G-IRM invariant regularizer.
We show \framework could improve performance of MTL models on benchmark datasets and could alleviate spurious correlation. 

\paragraph{Limitation Statement.}
Our analysis is based on label-label confounders. However, existing MTL datasets don't provide exact confounder changes to study spurious correlation problem. As mitigation, in analysis part, we create two synthetic datasets, and in experiment part, we adopt train/valid/test split with several attribution differences to mimic confounder changes. To further study spurious correlation in MTL, in the future, we'd like to construct benchmark MTL datasets with known confounder changes (or analyze how some key attribute changes lead to spurious correlation problem), build mathematical model based on it, and also explore and visualize which part of knowledge in real-world MTL datasets (e.g. Taskonomy) could be spuriously correlated to other tasks.



\paragraph{Acknowledgement.}

We sincerely thanks anonymous NeurIPS reviewers for their constructive comments and suggestions to improve this paper.
We thank Huan Gui, Kang Lee, Alexander D'Amour, Xuezhi Wang, Jilin Chen and Minmin Chen for insightful discussion and suggestion for this work. We also thank Ang Li for technical support for running experiments on Taskonomy, and Thanh Vu for running CityScape and NYUv2 experiments. Ziniu is supported by the Amazon Fellowship and Baidu PhD Fellowship.

\bibliography{causal}
\bibliographystyle{icml2021}

\section*{Checklist}
\begin{enumerate}

\item For all authors...
\begin{enumerate}
    \item Do the main claims made in the abstract and introduction accurately reflect the paper's contributions and scope?
    \answerYes{The main contribution of this paper is to point out the unique challenges of spurious correlation problem in MTL setup, and propose a workable solution. This has been discussed in abstract and intro.}
    \item Did you describe the limitations of your work?
    \answerYes{We've clearly stated the limitations and future directions to improve this work in Conclusion section.}
    \item Did you discuss any potential negative societal impacts of your work?
    \answerYes{We discuss how MTL could help solve label scarcity problem in related work. Thus, for some task with evil purpose, MTL could still help their performance, which has negative societal impacts.}
    \item Have you read the ethics review guidelines and ensured that your paper conforms to them?
    \answerYes{I've carefully read the guidances and ensured our paper conforms to them.}
\end{enumerate}

\item If you are including theoretical results...
\begin{enumerate}
    \item Did you state the full set of assumptions of all theoretical results?
    \answerYes{We state the assumption of proposition~\ref{pro:spur}. Specifically, we assume if MTL has spurious task correlation, then the model could utilize non-causal factors.}
    \item Did you include complete proofs of all theoretical results?
    \answerYes{The proof for Theorem~\ref{pro:spur} is described in Appendix~\ref{seq:proof}.}
\end{enumerate}

\item If you ran experiments...
\begin{enumerate}
    \item Did you include the code, data, and instructions needed to reproduce the main experimental results (either in the supplemental material or as a URL)?
    \answerYes{They are all included in the supplemental material.}
    \item Did you specify all the training details (e.g., data splits, hyperparameters, how they were chosen)?
    \answerYes{They are specified in Appendix~\ref{sec:data}.}
    \item Did you report error bars (e.g., with respect to the random seed after running experiments multiple times)?
    \answerYes{The detailed results with error bar is listed in Table~\ref{tab:res1}-\ref{tab:res5} in Appendix.}
    \item Did you include the total amount of compute and the type of resources used (e.g., type of GPUs, internal cluster, or cloud provider)?
    \answerYes{They are specified in Appendix~\ref{result}.}
\end{enumerate}

\item If you are using existing assets (e.g., code, data, models) or curating/releasing new assets...
\begin{enumerate}
    \item If your work uses existing assets, did you cite the creators?
    \answerYes{Yes, we've properly cite all used code and data in Appendix.}
    \item Did you mention the license of the assets?
    \answerYes{I cite the github link with license.}
    \item Did you include any new assets either in the supplemental material or as a URL?
    \answerYes{I'll submit our own code in supplemental material.}
    \item Did you discuss whether and how consent was obtained from people whose data you're using/curating?
    \answerNA{We do not use any personal data.}
    \item Did you discuss whether the data you are using/curating contains personally identifiable information or offensive content?
    \answerNA{We do not use any personal data.}
\end{enumerate}

\item If you used crowdsourcing or conducted research with human subjects...
\begin{enumerate}
    \item Did you include the full text of instructions given to participants and screenshots, if applicable?
    \answerNA{}
    \item Did you describe any potential participant risks, with links to Institutional Review Board (IRB) approvals, if applicable?
    \answerNA{}
    \item Did you include the estimated hourly wage paid to participants and the total amount spent on participant compensation?
    \answerNA{}
\end{enumerate}

\end{enumerate}

\newpage
\appendix

\section{Proof of Proposition \ref{pro:spur}}\label{seq:proof}

\subsection{Problem Definition}

\begin{wrapfigure}{r}{0.4\linewidth}
    \centering
    \includegraphics[width=0.95\linewidth]{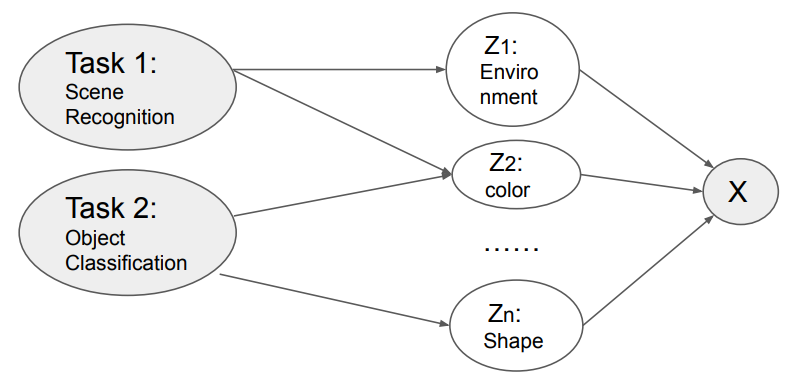}
    \caption{Illustrative Diagram of Causal Generative Model in MTL setting}
    \label{fig:generative}
\end{wrapfigure}


We consider two binary classification tasks, with $Y_a$ and $Y_b$ as variables from $\{ \pm 1\}$ for task label. The task labels are drawn from two different probabilities. For simplicity, we assume the probability to sample the two label value is balanced, i.e., $P(Y=1) = P(Y=-1) = 0.5$. Our conclusion could be extended to unbalanced distribution.

In this paper, we mainly study the spurious correlation between task labels. For simplicity, we define $P(Y_a = Y_b) = m_C, P(Y_a \neq Y_b) = 1-m_C$, where $m_C$ denotes that this correlation could change by different confounder $C_{dist}^{MTL}$. In some environments $m_C \neq 0.5$, meaning that the two tasks are correlated in these environments. To sum up, we could define the probability table as:
\begin{table}[h]
\centering
\begin{tabular}{c|c|c|}
\cline{2-3}
                              & $Y_a=1$  & $Y_a=0$ \\ \hline
\multicolumn{1}{|l|}{$Y_b=1$} &  $m_C$   &  $1-m_C$   \\ \hline
\multicolumn{1}{|l|}{$Y_b=0$} & $1-m_C$  &   $m_C$  \\ \hline
\end{tabular}
\caption{Probability table for $P(Y_a, Y_b)$, where $m_C$ denotes the correlation between the two task label.}
\end{table}

We consider two $d$-dimensional factors $F_a$ and $F_b$ representing the knowledge to tackle the two tasks. Both are drawn from Gaussian distribution:
\begin{align}
    F_a \sim \mathcal{N}(Y_a \cdot \mu_a, \sigma_a^2I), \ \ F_b \sim \mathcal{N}(Y_b \cdot \mu_b, \sigma_b^2I)
\end{align}
with $\mu_a, \mu_b \in \mathbb{R}^{d}$ denote the mean vectors and $\sigma_a$, $\sigma_b$ are covariance vectors. 

Our goal to learn two linear models $P(Y_{\{a/b\}}|F_a, F_b)=\text{sigmoid}(\beta F) = \text{sigmoid}(\beta_a F_a + \beta_b F_b)$.
We first consider the setting that we're given infinite samples. If we assume there's no traditional factor-label spurious correlation in single task learning, the bayes optimal classifier will only take each task's causal factor as feature, and assign zero weights to non-causal factors. The factor with the regression vector $\beta_a = \frac{\mu_a}{\sigma_a^2}$ for bayes optimal classifier of task $a$ and $\beta_b = 2\frac{\mu_b}{\sigma_b^2}$ for bayes optimal classifier of task $b$.

\subsection{Bayes Optimal Classifier for Multiple-Task} \label{bayes}
When we train a single model using both tasks, the optimal Bayes classifier will utilize the other non-causal factor due to the influence of spurious correlation quantified by $m_C$. To prove it, we take the first task with label $Y_a$ as an example and derive the optimal Bayes classifier as:
\begin{align}
    P(Y_a|F_a, F_b)& = \frac{P(Y_a,F_a, F_b)}{P(F_a, F_b)} = \frac{P(Y_a,F_a, F_b)}{\sum_{Y_a \in \{-1,1\}} P(Y_a, F_a, F_b)  }  \label{eq:formula}
\end{align}
while the probability of $P(Y_a, F_a, F_b)$ could be written as:
\begin{align}
&P(Y_a, F_a, F_b) = P(Y_a, F_a) \cdot P(F_b | Y_a, F_a)\\
    &= P(Y_a, F_a) \cdot P(F_b | Y_a)\\
    &= P(Y_a, F_a) \cdot \sum_{Y_b \in \{-1,1\}} P(F_b, Y_b | Y_a) \\
    &= P(Y_a) P(F_a|Y_a) \cdot \sum_{Y_b \in \{-1,1\}} P(F_b | Y_b) P(Y_b | Y_a)\\
    &\propto e^{Y_a \cdot F_a \beta_a } \cdot \big(  m_C  e^{Y_a \cdot F_b \beta_b} + (1-m_C)  e^{-Y_a \cdot F_b \beta_b} \big)\\
    &=m_C e^{Y_a (F_a\mu_a+F_b\mu_b)} + (1-m_C) e^{Y_a (F_a\mu_a-F_b\mu_b)}
\end{align}
By putting it back to equation(\ref{eq:formula}), we could get:
\begin{align}
    P(Y_a|F_a, F_b) = \frac{1}{1 + \frac{m_C e^{Y_a (F_a\beta_a+F_b\beta_b)} + (1-m) e^{Y_a (F_a\beta_a-F_b\beta_b)}}{m_C e^{-Y_a (F_a\beta_a+F_b\beta_b)} + (1-m) e^{-Y_a (F_a\beta_a-F_b\beta_b)}}} 
\end{align}
The formula shows that the optimal bayes classifier depends on the non-causal factor $F_b$ given $m_C \neq 0.5$.

To give two extreme, when $m_C = 1$:
\begin{align}
    P(Y_a|F_a, F_b) = \frac{1}{1 + e^{2Y_a (F_a\beta_a+F_b\beta_b)}}
\end{align}
In this way, the optimal classifier is $\beta = [2\beta_a, 2\beta_b]^T$ for the two factors $F_a$ and $F_b$.

When $m_C = 0.5$:
\begin{align}
    P(Y_a|F_a, F_b) = \frac{1}{1 + e^{2Y_a (F_a\beta_a)}}
\end{align}
In this way, the optimal classifier is $\beta = [2\beta_a, 0]^T$, which only utilizes the first factor $F_a$ and assign zero weights for the non-causal factor $F_b$.

\subsection{Classifier trained on limited dataset}
In the following we're considering the cases whether there's no task correlation in training set ($m_C = 0.5$). Though we have shown previously the optimal classifier should be invariant to non-causal factors given unlimited data, in reality with limited training dataset, the model could still utilize non-causal factors as noise.

Assume the training data contains spurious feature $S$ appended to causal feature $C$ for ground-truth linear model $Y=\theta^* C$, both under-parametrized and over-paramatrized linear model $\hat{Y} = \hat{\theta}C + \hat{\beta}S $ will assign non-zero weights $\hat{\beta}$ for spurious feature $S$.

Let $x \in \mathbb{R}^{(d+1) \times 1}$ denote the feature, where $x[1:d]=c$ is the causal feature, and $x[d+1]=s$ is spurious feature.

Let ground-truth linear model $y_i=f_{\theta^*}(x) = \theta^*  \cdot x_i + \epsilon_i = c \cdot \theta^*_c + \epsilon_i$, where $\theta^* = [\theta_c^*, 0] \in \mathbb{R}^{(d+1) \times 1}$ and $\epsilon \sim N(0, \sigma^2)$.

Given training dataset $X \in \mathbb{R}^{n \times (d+1)}$ and $Y = X \theta^* + \varepsilon = C \theta_c^* + \varepsilon \in \mathbb{R}^{n \times 1}$, the closed-form solution $\hat{\theta} \in \mathbb{R}^{(d+1) \times 1}$ for linear regression model  is:
\begin{align}
    \hat{\theta} = X^+ Y^+ = X^+ (X \theta^* + \varepsilon)
\end{align}
The generalization error is:
\begin{align}
    \mathcal{L} &= \mathbb{E}_{x} \Big[ \big(  (\hat{\theta} - \theta^*) \cdot x \big)^2 \Big]\\
    &= \mathbb{E}_{x} \Big[ \big(  (  X^+X - I ) \theta^* \cdot x + X^+\varepsilon \cdot x \big)^2 \Big] \\
    & = \mathbb{E}_{x} \Big[ \big( (  X^+X - I ) \theta^* \cdot x \big)^2\Big] + \sigma^2 \mathbb{E}_{x} \Big\lVert {(X^T)^+ x} \Big\rVert_2^2
\end{align}
The first term is bias and the second is variance.

If $X = [C, 0]$, which only contains causal feature without any spurious feature, we denote the learned parameter and loss as $\hat{\theta}_C$ and $\mathcal{L}_C$.

If $X = [C, S]$, which contains the spurious feature, we denote the learned parameter and loss as $\hat{\theta}_{S}$ and $\mathcal{L}_{S}$.

Our goal is to prove the learned parameter weight for the spurious feature is not zero. We'll study it in both underparamtrizied ($d+1 \leq n$) setting, where the solution is equivalent to least-square solution; and overparametrized ($d > n$), where the solution is equivalent to min-norm solution.

\subsubsection{Underparametrized Setting}
\paragraph{Loss}
Since $X \in \mathbb{R}^{n \times (d+1)}$ has independent column due to under parametrization assumption, we can find pseudo-inverse such that $X^+X = I$. Thus the bias term in $\mathcal{L}$ is 0, and we only need to consider the variance term.

\begin{align}
\mathcal{L}_S - \mathcal{L}_C = \sigma^2 \Big( \mathbb{E}_{x} \Big\lVert {
\begin{bmatrix}C^T\\S^T\end{bmatrix}^+ x} \Big\rVert_2^2 - \mathbb{E}_{x} \Big\lVert {\begin{bmatrix}C^T\\0\end{bmatrix}^+ x} \Big\rVert_2^2 \Big)
\end{align}

Since $||A^+x||_2^2 = \min_{Z:Az=x} ||z||_2^2$, and obviously $\Big\{ z \big | \begin{bmatrix}C^T\\S^T\end{bmatrix} z = x \Big\} \subseteq \Big\{ z \big | \begin{bmatrix}C^T\\0\end{bmatrix} z = x \Big\}$ as the first one has one more constraint. Therefore, $\Big\lVert {\begin{bmatrix}C^T\\S^T\end{bmatrix}^+ x} \Big\rVert \geq \Big\lVert {\begin{bmatrix}C^T\\0\end{bmatrix}^+ x} \Big\rVert_2^2$, and thus $\mathcal{L}_S \geq \mathcal{L}_C$.

\paragraph{weight}
 By the theorem 1 of \citep{BAKSALARY200716}, if $d+1 \leq n$, $X = [S, T] \in \mathbb{R}^{n \times (d+1)}$ has independent column, thus we have
\begin{align}
    X^+ = \begin{bmatrix}C^T\\S^T\end{bmatrix}^+ =  \begin{bmatrix}(I-Q)C(C^T(I-Q)C)^{-1} \\   
                  \frac{(I-P)S}{S^T(I-P)S}
    \end{bmatrix}
\end{align}
where $P=CC^T$, $Q=SS^T$.

Therefore,
\begin{align}
    \hat{\theta}_S[d+1] = \frac{(I-P)S}{S^T(I-P)S}Y = \frac{(I-P)S(C\theta^*_C + \varepsilon)}{S^T(I-P)S}
\end{align}

\subsubsection{Overparametrized Setting}
In this setting the closed-form solution is equivalent to minimum-norm solution, such that:
\begin{align}
    & \hat{\theta} = \arg \min_{\theta} ||\theta||_2^2 \\
    & s.t. \ \ X \theta = Y
\end{align}
\paragraph{weight}
Since $X$ is have full row rank, $(XX^T)^-1$ exists, thus we have:
\begin{align}
    X^+ = X^T (XX^T)^{-1}
\end{align}
Based on the Sherman-Morrison formula, we have:
\begin{align}
    (XX^T)^{-1} = (CC^T + SS^T)^{-1} = G-\frac{GSS^TG}{1+S^TGS}
\end{align}
where $G=(CC^T)^{-1}, u=\frac{b^TG}{1+b^TGb}$. Therefore:
\begin{align}
    X^+ = \begin{bmatrix}C^T\\S^T\end{bmatrix}^+ = \begin{bmatrix}(I-bu)C^+\\ u\end{bmatrix}
\end{align}
Thus \begin{align}
    \hat{\theta}_S[d+1] = \frac{b^TG}{1+b^TGb}Y = \frac{b^TG(C\theta^*_C + \varepsilon)}{1+b^TGb}
\end{align}

To sum up, given limited training dataset, even without spurious correlation between tasks, and non-causal features only serve as noise, the model could still learn to assign non-zero weights to non-causal features to overfit the dataset. Therefore, in MTL setting, when the number of tasks increase, the shared representation encodes many causal features from different tasks. Even without spurious correlation, it will lead to overfitting issue. And such problem could be exacerbated by spurious correlation that we show in section~\ref{bayes}.

\newpage

\section{Synthetic Analysis of Multi-SEM with more tasks and saliency map}\label{sec:more}

\begin{wrapfigure}{r}{0.5\linewidth}
    \centering
    \includegraphics[width=1.0\linewidth]{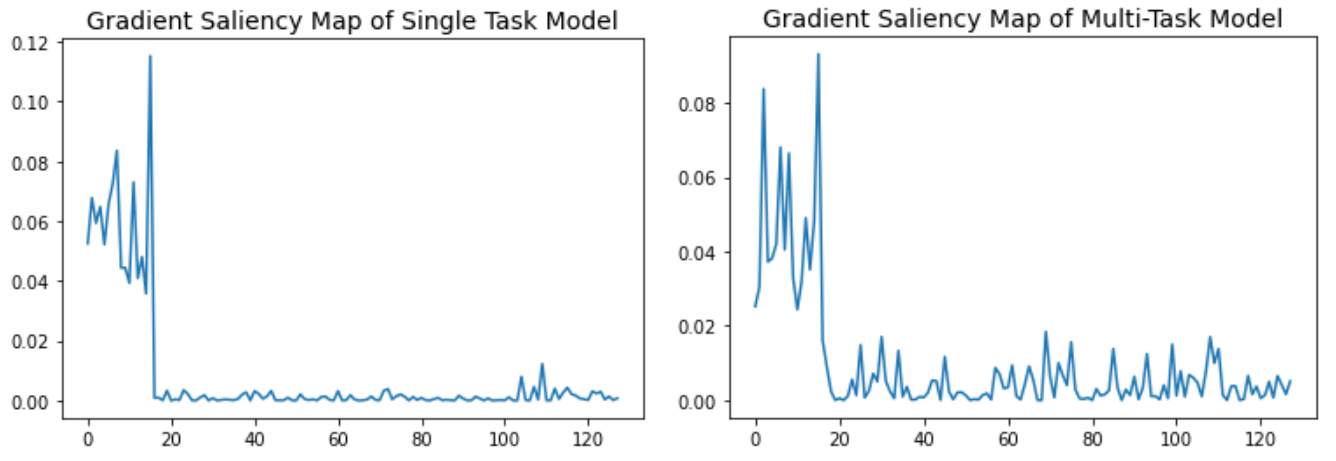}
    \caption{The gradient saliency map of Multi-SEM. The model trained by MTL exploits non-causal features (spurious) more.}
    \label{fig:sem}
\end{wrapfigure}

In section~\ref{sec:analysis} we compare model trained by MTL with STL with two tasks. Here we show the results conducted in Multi-SEM with more than two tasks in Table~\ref{tab:more}. The results show decreasing Acc$_{val}$ and higher usage of spurious feature $\rho_{spur}$ compared with STL, with increasing number of tasks. This matches our hypothesis that MTL could incorporate more non-causal features / factors into shared representation, increasing the risk of utilizing overfitting.
We also show the saliency map for each feature dimension in Figure~\ref{fig:sem}. It shows that the model trained by MTL exploits non-causal features (dimension 20-120) more than the model trained by STL.
All these results empirically support our claim that with spurious task correlation, model trained by MTL utilize non-causal factors more and generalize worse than STL.

\begin{table}[h]
\centering
\begin{tabular}{ll|ccccccc} \toprule
\multicolumn{2}{c|}{$\#\text{Tasks}$}    & 2 & 3 & 4 & 5 &6 &7 &8\\ \midrule
\multirow{2}{*}{MTL} & Acc$_{val}$ & 0.846 & 0.838 & 0.824 & 0.809 & 0.785 & 0.752 & 0.719\\
~ & $\rho_{spur}$ & 0.328 & 0.357 &  0.391 & 0.429 & 0.475 & 0.530 & 0.594\\ \midrule 
\multirow{2}{*}{STL} & Acc$_{val}$ & 0.874 & 0.861 & 0.848 & 0.836 & 0.827 & 0.810 & 0.797\\
~ & $\rho_{spur}$ & 0.261 & 0.289 & 0.314 & 0.354 & 0.385 & 0.407 & 0.435\\
\bottomrule
\caption{Results on Multi-SEM with more than 2 tasks.}\label{tab:more}
\end{tabular}
\end{table}

\section{Pseudo-Code and more discussion of \framework}\label{sec:code}

\begin{algorithm}[t!]
    \caption{Pseudo-Code of proposed MT-CRL (use $\mathcal{L}_{G\text{-}IRM}^{Var}$ as invariant regularizer)}
    \label{alg:train-extend}
\begin{algorithmic}[1]
    \REQUIRE shared encoders with $K$ different neural modules $\Phi = \big[\Phi_i(\cdot)\big]_{i=1}^K$, biadjacency matrix $A=\text{sigmoid}(\theta) \in [0,1]^{T \times K}$, per-task predictors $\mathcal{F} = \{ f_t\}_{t\in \mathcal{T}}$, minibatch with environment label and loss function for each task $\mathcal{B}_{t\in \mathcal{T}} = \{X_t, Y_t, E_t, \mathcal{L}_{t} \}_{t\in \mathcal{T}}$
    \STATE $\mathcal{L}_{\mathcal{B}} = 0$
    \FOR{each task $t \in \mathcal{T}$} 
        \STATE Get $X_t, Y_t, \mathcal{L}_{t}$ from $\mathcal{B}_t$
        \STATE $\mathbf{Z} = \big[\mathbf{Z}_i\big]_{i=1}^K = \big[\Phi_i(X_t)\big]_{i=1}^K$ 
        \STATE $\hat{Y}_t(\mathbf{X_t}) = f_t\big(\sum_{i}A_{t,i} \cdot \mathbf{Z}_i)\big) = f_t\big(\sum_{i}A_{t,i} \cdot \Phi_i(\mathbf{X_t})\big)$
        \STATE $\mathcal{L}_{\mathcal{B}} = \mathcal{L}_{\mathcal{B}} + R_t\big(\Phi, A_t, f_t \big) = \mathcal{L}_{\mathcal{B}} +  \mathcal{L}_{t}(\hat{Y}_t(\mathbf{X_t}), Y_t)$
        \STATE $\mathcal{L}_{\mathcal{B}} = \mathcal{L}_{\mathcal{B}} +  \mathcal{L}_{decor}(\Phi)_t =  \mathcal{L}_{\mathcal{B}} + \lambda_{decor} \cdot {\sum_{i = 1}^{k}\sum_{j = i+1}^k}\big\lVert \rho\big(\Phi_i(X_t), \Phi_j(X_t)\big)  \big\rVert_F^2$
    \ENDFOR 
    \STATE $\mathcal{L}_{\mathcal{B}} = \mathcal{L}_{\mathcal{B}} + \mathcal{L}_{graph}(A) =  \mathcal{L}_{\mathcal{B}} + \Big(\lambda_{sps} \cdot ||A||_1 - \lambda_{bal} \cdot \text{Entropy}\big(\frac{\sum_t A_{t, *}}{\sum_{t, i} A_{t, i}}\big)\Big)$
    \STATE $grad = \nabla_{A, \mathcal{F},\Phi }\ \mathcal{L}_{\mathcal{B}}$
    \STATE Detach $\mathcal{F} = \{f\}_t$ from computational graph (use $tf.stop\_gradient$ or $torch.zero\_grad$)
    \STATE $\mathcal{L}_{G\text{-}IRM}^{Var}(\Phi,  A | f) = 0$
    \FOR{each task $t \in \mathcal{T}$} 
        \STATE Get environment label $E_t$. In our experimental setting it's train and valid set.
        \STATE $\mathcal{L}_{G\text{-}IRM}^{Var} = \mathcal{L}_{G\text{-}IRM}^{Var} + \sum_{e \in E_t} \frac{1}{|E_t|} \Big\lVert \nabla_{A = A_t} R^e_t\big(\Phi, A, f_t \big) - \text{Avg}_e\Big(\nabla_{A = A_t} R^e_t\Big) \Big\rVert^2$
    \ENDFOR
    \STATE $grad = grad + \nabla_{A, \Phi}\ \mathcal{L}_{G\text{-}IRM}^{Var}(\Phi,  A | f)$
    \STATE Use optimizer to update the model via gradient $grad$

\normalsize
\end{algorithmic}
\end{algorithm}

The full psudo-code of proposed MTL is shown in Alg.~\ref{alg:train-extend}. We first use disentangled MMoE model to calculate loss for each task $R_t\big(\Phi, A_t, f_t \big)$, and also calculate disentangled and graph regularization. We then calculate invariant regularization over train/valid split. The most important part is line 11 we detach the per-task predictors from computational graph, so that when we calculate gradient (via $loss.backward$), we only calculate gradient over graph $A$ and encoder $\Phi$.

Ideally the invariant loss should be calculated based on different environmental split, similar to what is utilized in existing Out-Of-Distribution Generalization works. However, in MTL setting, there's no datasets designed specifically for studying OOD generalization or spurious correlation. To make current approach suitable for real-world applications, we only utilize two environment split (i.e. train and valid from existing datasets).

\begin{wrapfigure}{r}{0.33\linewidth}
    \centering
    \includegraphics[width=0.9\linewidth]{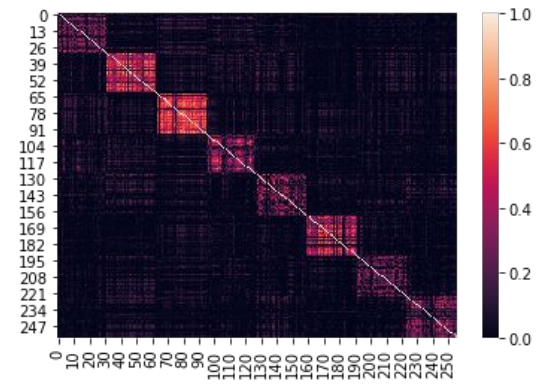}
    \caption{The heatmap of mutual correlation $\rho(\mathbf{Z}_i, \mathbf{Z}_j)$ between every pairs of modules.}
    \label{fig:heatmap}
\end{wrapfigure}

Noted that in our framework we adopt a simple linear correlation regularization to enforce disentanglement. This regularization only forces representation to be linearly de-correlated, and a more strict solution might be reducing the mutual information (MI). However, existing methods to minimizing MI requires either knowing the latent distribution (e.g. InfoGAN. We report BetaVAE in Table 3 with similar intuition but performs worse) or over estimated MI (e.g. MINE). We indeed tried adding discriminator for every module pair and adopted Minmax training to minimize estimated MINE. The result is unstable and no better. Module output's norm is very large and only the centers are seperated rather than disentangled. Therefore, we only utilize the linear de-correlation methods that perform well in our experiments. We show the mutual correlation of every pairs of modules in Figure~\ref{fig:heatmap} learned in MultiMNIST dataset. It shows that after learning, the modules indeed learn to be linearly de-correlated between each other, and only have correlated neurons within each module.

\section{Details about Dataset}\label{sec:data}

\subsection{Synthetic Datasets}\label{sec:syntheticdataset}

\paragraph{Multi-SEM.}
We mostly follow the setting of linear Structural Equation Model (SEM) proposed by \citet{DBLP:conf/iclr/RosenfeldRR21}. The two binary-classification task labels $Y_a$ and $Y_b$ are causally related to two distinctive factors $F_a$ and $F_b$ respectively via Gaussian distribution. 
We define the spurious correlation of the two labels by the probability that the two labels are the same: $C^{MTL}_{dist} = P(Y_a = Y_b)$. We set different $C_{dist}$ for training and test sets to simulate distribution shifts.

\begin{figure}
    \centering
    \includegraphics[width=0.8\textwidth]{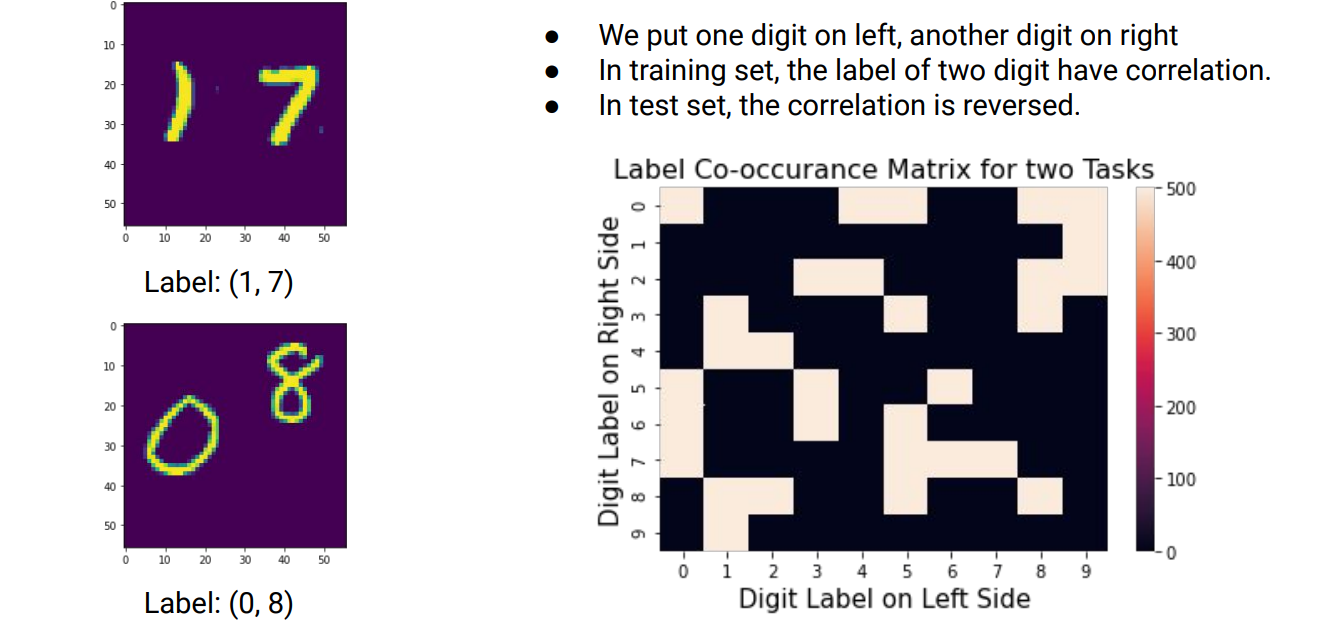}
    \caption{Illustrative figure for spurious Multi-MNIST dataset used for analaysis.}
    \label{fig:mnist}
\end{figure}

\paragraph{Multi-MNIST.}
We modified the multi-digit MNIST~\citep{mulitdigitmnist}, which samples two digit pictures and put in left and right position. The generative variables $F_{left}, F_{right}$ are the digit images and data input is simply their concatenation: $X = [F_{left}, F_{right}]$. We define the task correlation $C^{MTL}_{dist}$ by co-occurrence probability of the two digit labels. We randomly shuffle the label pairs and split the training and test set such that the class label pairs do not overlap. An illustrative data point and the label pairs in training set is shown in Figure~\ref{fig:mnist}.

\subsection{Real-world Datasets}\label{sec:dataset}

\textbf{Multi-MNIST}~\citep{DBLP:journals/tiis/HarperK16} is a multi-task variant of MNIST dataset, which samples two digit pictures and put in left and right position. We mainly modified from the this code repo\footnote{\url{https://github.com/shaohua0116/MultiDigitMNIST}} to generate the dataset. We sample 10,000 images for each label pair, so totally there are 1M data samples. As discussed in analysis section, to mimic distribution shifts (i.e., task correlation $C_{dist}^{MTL}$), we randomly shuffle the label pairs and split the train, valid and test set with ratio 3:1:1, such that every image co-occurrence correlation will no longer appear again in test set.  
We utilize the same CNN architectures and hyperparameter adopted in~\citet{DBLP:conf/nips/YuK0LHF20} as base encoder, and one-layer MLP as per-task predictor.

\textbf{MovieLens}~\citep{DBLP:journals/tiis/HarperK16} is a Movie recommendation dataset that
contains 10M rating records\footnote{\url{https://files.grouplens.org/datasets/movielens/ml-10m.zip}} of 10,681 movies by 71,567 users from Jan. 1996 to Dec. 2008. We consider the rating regression for movies in each genre as different tasks. There are totally 18 different genres, including Action, Adventure, Animation, Children's, Comedy, Crime, Documentary, Drama, Fantasy, Film-Noir, Horror, Musical, Mystery, Romance, Sci-Fi, Thriller, War and Western. To mimic distribution shifts across train, valid and test set, we split the data based on timestamp with ratio 8:1:1, and filter out non-overlapping users and movies from each set.
We utilize a embedding layer followed by two-layer MLP as base encoder, and one-layer MLP as per-task predictor.

\textbf{Taskonomy}~\citep{DBLP:conf/cvpr/ZamirSSGMS18} is a large-scale MTL benchmark dataset of indoor scene images from various buildings\footnote{\url{http://taskonomy.stanford.edu/}}. Every image has annotations for a set of diverse computer vision tasks. We follow the setting of~\citep{DBLP:journals/corr/abs-2010-02418} to use 8 tasks, including curvature estimation, object classification, scene classification, surface normal estimation, semantic segmentation, depth estimation, occlusion edge, 2D keypoint estimation and 3D keypoint estimation. For these tasks, object and scene classification tasks are trained using cross entropy loss, semantic segmentation using pixelwise cross entropy, curvature estimation using L1 loss, and all other tasks using L2 loss.
To mimic distribution shift, we select images from non-overlapping 48, 3, 3 buildings as train, valid and test set. The total training size is 324864 samples.
We use Resnet-50 model as our base encoder network, and 15-layer CNN model with upsampling blocks as the per-task predictor.

\textbf{NYUv2}~\citep{DBLP:conf/eccv/SilbermanHKF12} is a dataset of 1449 RGB-D indoor scene images\footnote{\url{https://cs.nyu.edu/~silberman/datasets/nyu_depth_v2.html}} with three tasks: 13-class semantic segmentation, depth estimation, and surface normals prediction. We use mean Intersection-Over-Union (mIoU), Relative Error (Rel Err) and Angle Distance as evaluation metric for the three tasks respectively. To mimic distribution shift, we split the dataset by scene labels into train, valid and test set with ratio 8:1:1.
We follow the setting adopted in~\citet{DBLP:conf/nips/YuK0LHF20} to use Segnet~\citep{DBLP:journals/corr/BadrinarayananK15} as the base encoder.

\textbf{CityScape}~\citep{DBLP:conf/cvpr/CordtsORREBFRS16} is a dataset of street-view images\footnote{\url{https://www.cityscapes-dataset.com/}} with two tasks: semantic segmentation and depth estimation. We use mean Intersection-Over-Union (mIoU) and Relative Error (Rel Err) as evaluation metric for the three tasks respectively. We follow the same data pre-processing procedure of the original paper, and split images based on city into 2475, 500 and 500 train, valid and test samples. We follow the setting adopted in~\citet{DBLP:conf/nips/YuK0LHF20} to use Segnet~\citep{DBLP:journals/corr/BadrinarayananK15} as the base encoder.





\begin{table*}[ht!]
\centering
\hspace*{-.2in}
\scriptsize
\begin{tabular}{l|cccccc|c} \toprule
 \textbf{Tasks (Metric)}    & \textbf{STL} & \textbf{MTL}  & \textbf{PCGrad}  & \textbf{GradVac}  & \textbf{DANN}  & \textbf{IRM}  & \textbf{MT-CRL + $\mathcal{L}_{G\text{-}IRM}^{Var}$}\\ \midrule
 Left-Digit (Acc.) & 0.871 $\pm$ 0.018 & 0.844 $\pm$ 0.019 & 0.880 $\pm$ 0.019 & 0.884 $\pm$ 0.017 & 0.878 $\pm$ 0.020 & 0.887 $\pm$ 0.010 & 0.912 $\pm$ 0.018\\ 
Right-Digit (Acc.) & 0.877 $\pm$ 0.015 & 0.848 $\pm$ 0.018 & 0.888 $\pm$ 0.017 & 0.886 $\pm$ 0.018 & 0.884 $\pm$ 0.017 & 0.889 $\pm$ 0.015 & 0.918 $\pm$ 0.019\\ \bottomrule
\end{tabular}
\caption{Results for Multi-MNIST dataset. }\label{tab:res1}
\end{table*}

\begin{table*}[ht!]
\centering
\hspace*{-.2in}
\scriptsize
\begin{tabular}{l|cccccc|c} \toprule
 \textbf{Metric}    & \textbf{STL} & \textbf{MTL}  & \textbf{PCGrad}  & \textbf{GradVac}  & \textbf{DANN}  & \textbf{IRM}  & \textbf{MT-CRL + $\mathcal{L}_{G\text{-}IRM}^{Var}$}\\ \midrule
 Avg. MSE & 0.894 $\pm$ 0.006 & 0.892 $\pm$ 0.005 & 0.892 $\pm$ 0.006 & 0.891 $\pm$ 0.005 & 0.890 $\pm$ 0.007 & 0.890 $\pm$ 0.004 & 0.884 $\pm$ 0.006\\  \bottomrule
\end{tabular}
\caption{Results for MovieLens dataset. }\label{tab:res2}
\end{table*}

\begin{table*}[ht!]
\centering
\scriptsize
\begin{tabular}{l|cccccc|c} \toprule
 \textbf{Tasks (Metric)}    & \textbf{STL} & \textbf{MTL}  & \textbf{PCGrad}  & \textbf{GradVac}  & \textbf{DANN}  & \textbf{IRM}  & \textbf{MT-CRL + $\mathcal{L}_{G\text{-}IRM}^{Var}$}\\ \midrule
object classification (Cross Entropy)& 3.37 & 3.18 & 3.09 & 3.06 & 3.13 & 3.16 & 3.01\\ 
scene classification (Cross Entropy) & 2.65 & 2.59 & 2.54 & 2.51 & 2.58 & 2.59 & 2.47\\ 
semantic segmentation  (Cross Entropy)  & 1.68 & 1.54 & 1.47 & 1.49 & 1.53 & 1.56 & 1.43\\   
curvature estimation (L1 Loss) & 0.246  & 0.224  & 0.218  & 0.212  & 0.237  & 0.226  & 0.208\\  
surface normal estimation (L2 Loss)  & 0.138 & 0.141 & 0.136 & 0.139 & 0.152 & 0.150 & 0.125\\ 
occlusion edge detection (L2 Loss) & 0.134 & 0.138 & 0.132 & 0.133 & 0.137 & 0.141 & 0.128\\ 
2D keypoint estimation  (L2 Loss) & 0.176 & 0.171 & 0.167 & 0.163 & 0.169 & 0.168 & 0.158\\ 
3D keypoint estimation  (L2 Loss) & 0.194 & 0.205 & 0.199 & 0.196 & 0.204 & 0.201 & 0.191\\ \bottomrule
\end{tabular} 
\caption{Results for Taskonomy dataset. }\label{tab:res3}
\end{table*}

\begin{table*}[ht!]
\centering
\scriptsize
\begin{tabular}{l|cccccc|c} \toprule
 \textbf{Tasks (Metric)}    & \textbf{STL} & \textbf{MTL}  & \textbf{PCGrad}  & \textbf{GradVac}  & \textbf{DANN}  & \textbf{IRM}  & \textbf{MT-CRL + $\mathcal{L}_{G\text{-}IRM}^{Var}$}\\ \midrule
Segmentation (mIoU) & 13.27 & 17.64 & 19.64 & 19.68 & 17.12 & 17.54 & 19.81\\ 
Depth (Rel Err)  & 0.653 & 0.651 & 0.591 & 0.593 & 0.637 & 0.639 & 0.585\\ 
Surface Normal (Angle Distance)  & 35.18 & 31.52 & 30.98 & 31.04 & 31.69 & 32.03 & 30.85 \\ \bottomrule
\end{tabular}
\caption{Results for NYU-V2 dataset. }\label{tab:res4}
\end{table*}

\begin{table*}[ht!]
\centering
\small
\begin{tabular}{l|cccccc|c} \toprule
 \textbf{Tasks (Metric)}    & \textbf{STL} & \textbf{MTL}  & \textbf{PCGrad}  & \textbf{GradVac}  & \textbf{DANN}  & \textbf{IRM}  & \textbf{MT-CRL + $\mathcal{L}_{G\text{-}IRM}^{Var}$}\\ \midrule
Segmentation (mIoU) & 50.87 & 51.63 & 52.84 & 52.76 & 51.91 & 52.05 & 53.12\\ 
Depth (Rel Err)  & 33.85 & 32.75 & 32.12 & 32.08 & 32.71 & 32.64 & 31.86\\ \bottomrule
\end{tabular}
\caption{Results for CityScape dataset. }\label{tab:res5}
\end{table*}

\begin{table*}[ht!]
\centering
\begin{tabular}{l|cccccccc} \toprule
 $K$    & 1 & 2 & 4 & \textbf{8} & 16 & 32 & 64 & 128\\
Acc. & 0.824 & 0.897 & 0.904 & \textbf{0.915} & 0.911 & 0.902 & 0.893 & 0.882\\
\bottomrule
\end{tabular}
\caption{Hyperparameter tuning results for number of module ($K$) over Multi-MNIST dataset. }\label{tab:hyper}
\end{table*}

The results show that a middle number of module ($K$=8) achieves the best performance under the same size of model. Note that disentangled representation learning methods like BetaVAE assume that every dimension is mutually independent ($K=128$ in our case), which restricts the model capacity. Therefore, a middle K is a trade-off between model disentanglement and capacity. In all other datasets, we just use $K=8$ by default and didn't do further tuning.

\section{Detailed Results on each Dataset}\label{result}
We report the performance on each task for the five benchmark datasets in Table~\ref{tab:res1}-\ref{tab:res5}. We use 8 GPU to run each experiments. As shown in the tables, the scale of different task's evaluation metric differ a lot, and thus in the main paper we adopt relateive performance improvement compared to vanilla MTL to evaluate each method. Nevertheless, our \framework with invariance regularization could achieve the best results over nearly all the tasks.

\section{Determine the number of modules $K$}\label{hyper}
$K$ is a hyperparameter that could be tuned. To control the same model complexity, we fix the output dimension $d$, i.e., 128, and then each module's dimension is $\frac{d}{K}$. We show the results on Multi-MNIST with different $K$ in Table~\ref{tab:hyper}.

The results show that a middle number of module (K=8) achieves the best performance under the same size of model. In all other datasets, we just use $K=8$ by default and didn't do further tuning to test our method's generalization capacity, this could avoid the situation that the final performance improvement is mainly caused by extensive hyper-parameter tuning.

\paragraph{Capacity-Disentanglement Tradeoff}

Noted that with a fixed number of dimension $d$, with larger $K$, the model capacity is reduced. The widely adopted disentangled representation learning methods like BetaVAE mostly assume that every dimension is mutually independent ($K=d$ in our case), which restricts the model capacity to extreme case. And the results in Table~\ref{tab:ablation} also show that our current disentangled approach performs empirically better than BetaVAE. One potential reason is that we choose allow a middle K is a trade-off between model disentanglement and capacity, in which only the dimension across the block is de-correlated, why the ones within block could still correlated, so as to maintain model expressiveness. 

Note that the optimal choice of $K$ should ideally should be proportional to number of true generative factors that are related to downstream tasks. Therefore, for dataset with a large amount of tasks, we should choose larger number of $K$, and also consider increasing the total number of dimension $d$ to increase the model capacity while maintaining disentanglement. In our paper for large dataset such as Taskonnomy we didn't do further tuning due to limited resources, so the performance could be potentially further improved, which we leave for future exploration.

\section{More Case Studies to show \framework can alleviate spurious correlation}\label{sec:case}

Here we show more details about the case study we conduct for analyzing how \framework could alleviate spurious correlation, as a complementary.

As introduced in case study, we use the task-to-module gradients $\frac{\partial( f(\Phi(x))[y])}{\partial F}$ to illustrate how each task utilize each module. We could utilize the (valid-train) score to show which module is used by training set but not helpful for valid set, meaning it is spurious.

\begin{figure}[t]
    \centering
    \includegraphics[width=1\textwidth]{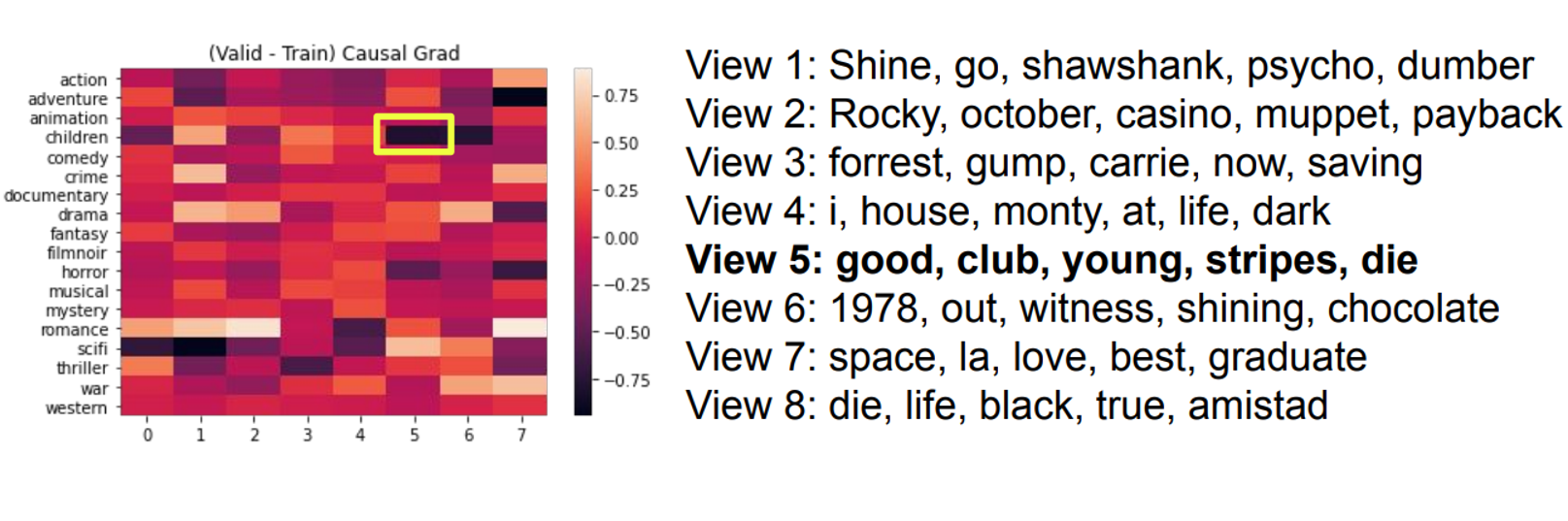}
    \caption{(valid-train) Task-to-Module gradients of model \textbf{without} MT-CRL on MovieLens.}
    \label{fig:movie_before}
\end{figure}

We first show detailed results on MovieLens dataset. As shown in Figure~\ref{fig:movie_before}, without \framework, there eixst many modules assigned negative (valid-train) causal grad, as shown in the color bar. Among them, view 5 is mostly inconsistent with the training results as we show in case study. Also, the other modules' key words are also not very accurate to describe the properties of each movie.

\begin{wrapfigure}{r}{0.45\linewidth}
    \centering
    \includegraphics[width=0.9\linewidth]{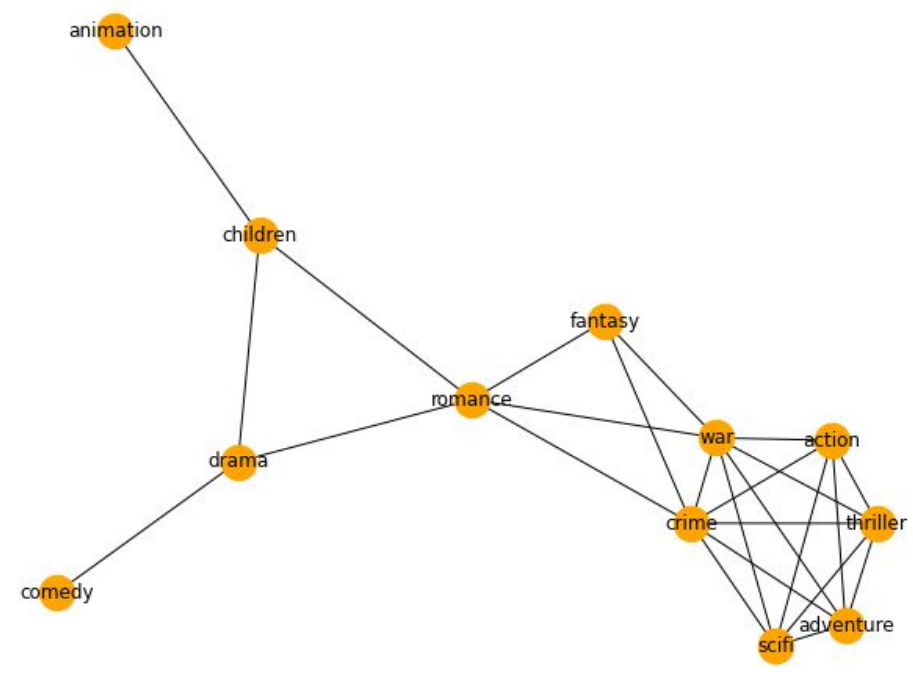}
    \caption{The task similarity induced by causal graph $A$ for MovieLens dataset (threshold = 0.1).}
    \label{fig:similarity}
\end{wrapfigure}

\begin{figure}[t]
    \centering
    \includegraphics[width=1\textwidth]{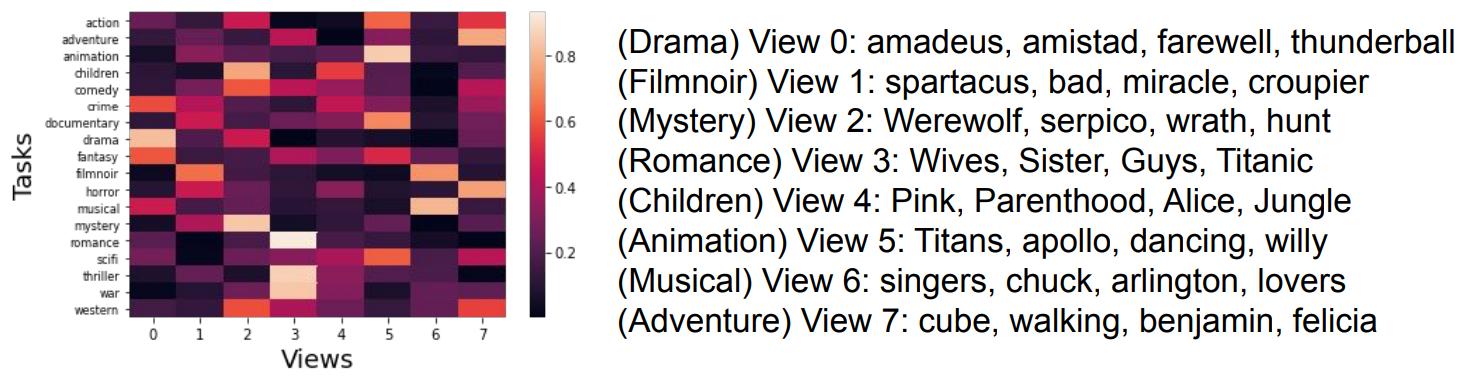}
    \caption{(valid-train) Task-to-Module gradients of model \textbf{with} MT-CRL on MovieLens.}
    \label{fig:movie_after}
\end{figure}

After we add \framework, as shown in Figure~\ref{fig:movie_after}, all of the modules receive positive (valid-train) causal grad, meaning that they either not utilized in training stage, or every used modules are still useful in valid stage. In addition, all the modules' key words are much more accurate to describe each type of movie than before.

In addition, our learned task biadjacency graph $A$ could also be used to describe the similarity between task. If two tasks share more causal feature, they are more similar. We thus cauculate the task-averaged score of $A$, and plot a sparse smilarity graph in Figure~\ref{fig:similarity}. It shows that our \framework could learn to group similar types of adult movies, such as war, crime, thriller, adventure into the same group in the right down part, and link children-friendly movies, such as animation and children together. romance movie is a link between adult cluster and children movie. This similarity graph matches our human expectation, showing that our learned causal graph indeed help similar group use similar causal features.

\begin{figure}[t]
    \centering
    \includegraphics[width=1\textwidth]{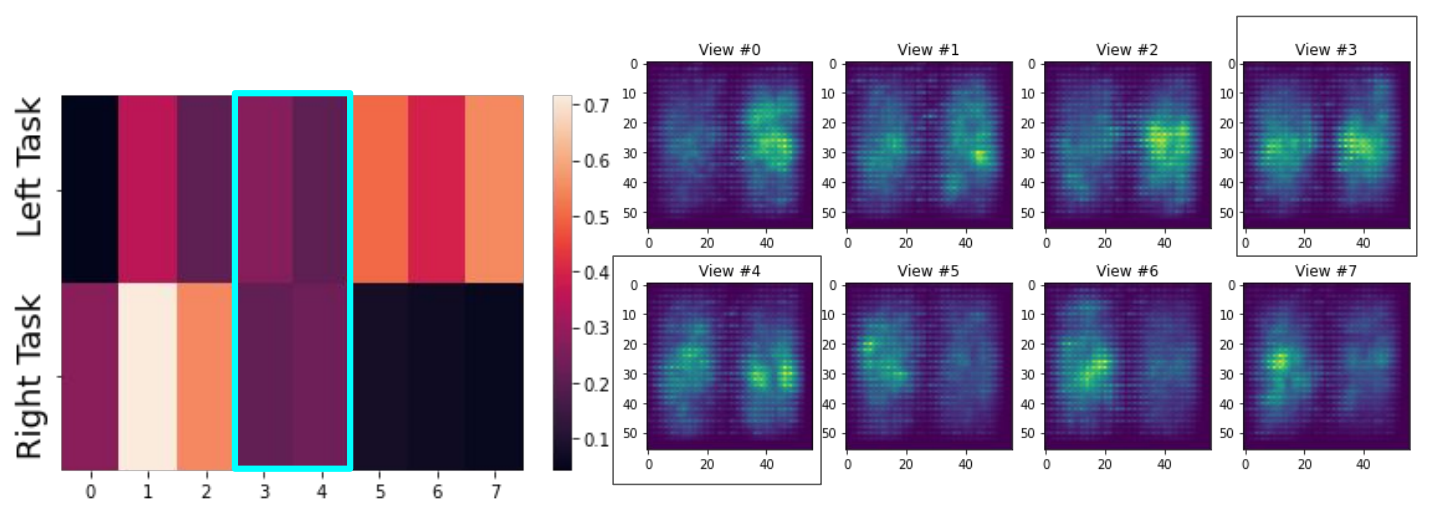}
    \caption{(valid-train) Task-to-Module gradients of model \textbf{without} MT-CRL on Multi-MNIST.}
    \label{fig:mnist_before}
\end{figure}

\begin{figure}[t]
    \centering
    \includegraphics[width=1\textwidth]{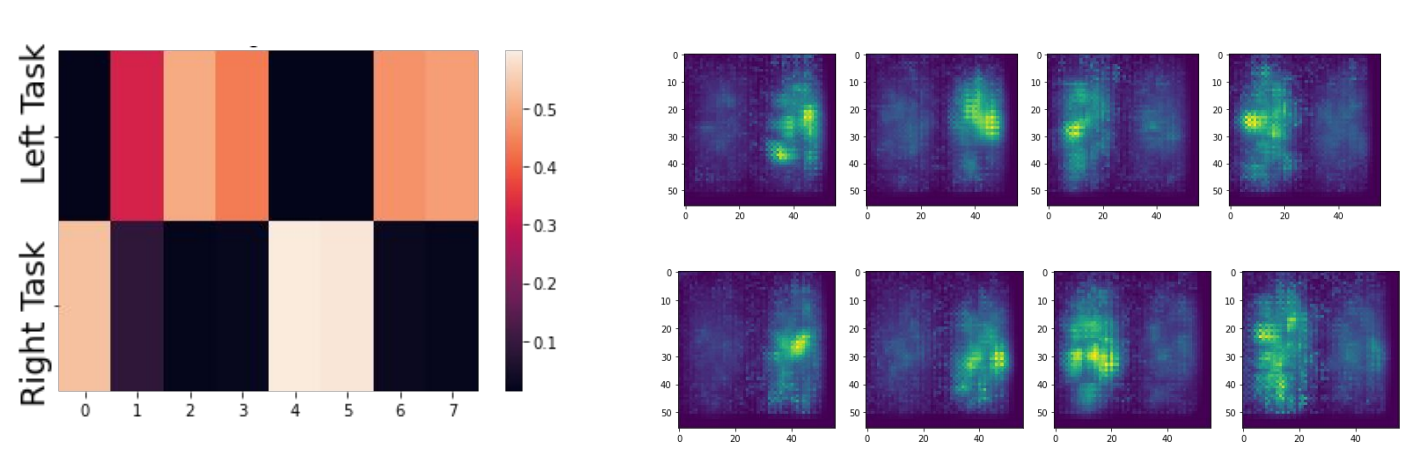}
    \caption{(valid-train) Task-to-Module gradients of model \textbf{with} MT-CRL on Multi-MNIST.}
    \label{fig:mnist_after}
\end{figure}

\begin{wrapfigure}{r}{0.4\linewidth}
    \centering
    \includegraphics[width=1.0\textwidth]{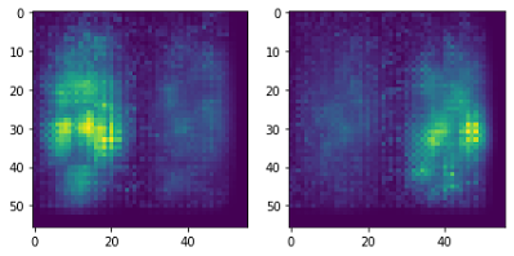}
    \caption{The gradient saliency map of left-digit and right-digit classifier trained via \framework. Compared with Figure~\ref{fig:vis}(b), \framework indeed helps alleviate spurious correlation.}
    \label{fig:mnistall}
\end{wrapfigure}

We then show the (valid-train) Task-to-Module gradients over Multi-MNIST datasets. It is very apparent that the two digit classifier doesn't share any overlapping causal features. However, as shown in Figure~\ref{fig:mnist_before}, without \framework, the model still learns to assign similar weights to module 3 and 4. This is also illustrated by the gradient saliency map for each module. Module 3 and 4 have high attention on both left and right side.

With \framework, in Figure~\ref{fig:mnist_after}, the learned task-to-module assignment is much sparse and clear. Also each module's saliency map only focus on one side of pixels. By looking at each task output's saliency map, in Figure~\ref{fig:mnistall}, we can see the model with \framework can help to focus only on causal part, compared with Figure~\ref{fig:vis}(b) that have high weights on both.

Both the MovieLens and Multi-MNIST case studies show that \framework could help alleviate spurious correlaiton issue. For the other datasets, such as Taskonomy, NYUv2 and CityScape, their task output layer is very different and thus it's hard to get normalized gradient to show in one figure. In the future, we plan to do more thorough analysis by manually label several spurious feature or environmental groups, and design better methods to visualize how MTL model utilize non-causal features.

\begin{figure}[t!]
\begin{floatrow}
\ffigbox{%
  \includegraphics[width=1\columnwidth]{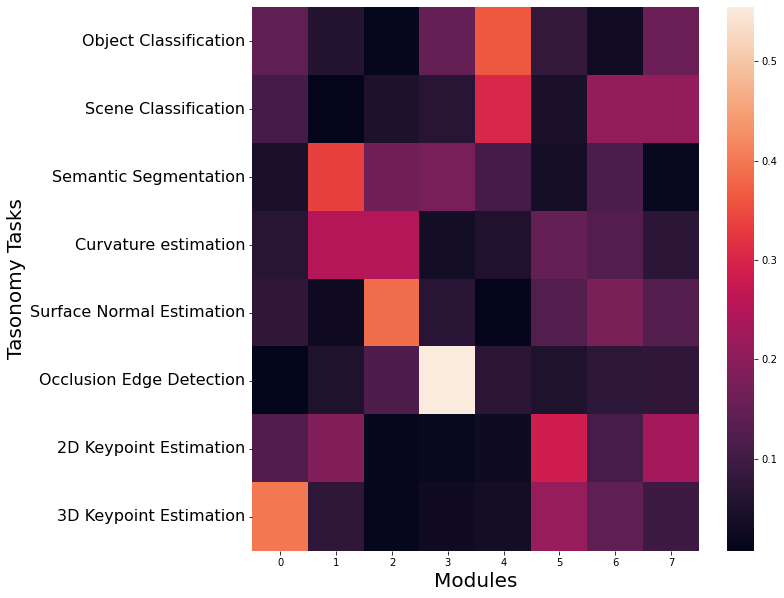}
}{%
  \caption{Task-to-Module Routing Graph ($A$) of model trained on Taskonomy dataset.}
    \label{fig:task1}
}
\ffigbox{
    \includegraphics[width=1.1\columnwidth]{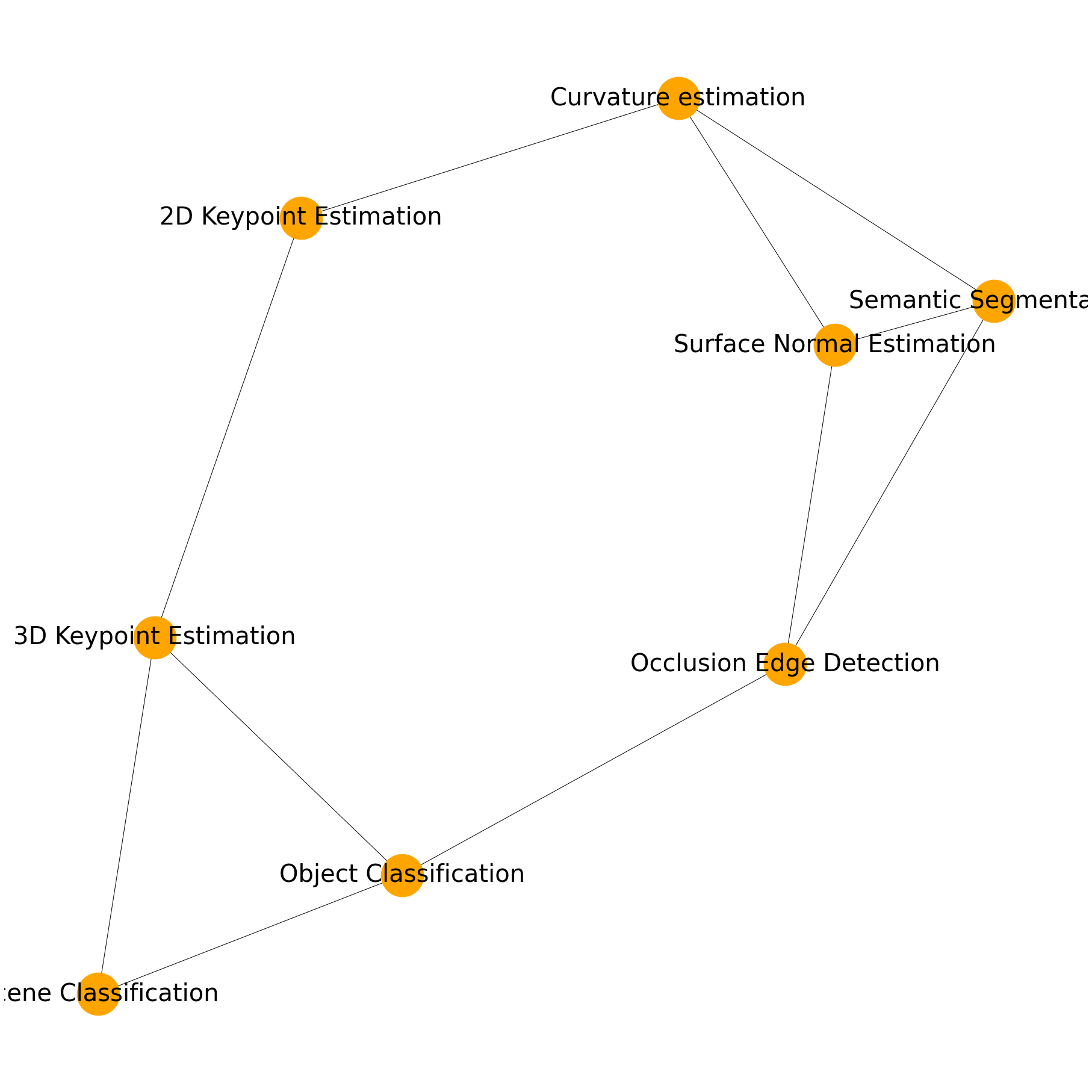}
}{%
   \caption{Task Similarity graph induced by Task-to-Module Graph $A$ for Taskonomy dataset (threshold = 0.1).}
    \label{fig:task2}
}
\end{floatrow}
\end{figure}

\paragraph{Case Study on Taskonomy.}
In addition, we show the Task-to-Module routing graph and also the induced task similarity graph of Tasknomy dataset. As is shown in the figure, some similar task like 2D keypoint Estimation and 3D keypoint are liked together, and also the hard task like semantic segmentation receives information from curvature estimation, surface normal estimation and occlusion edge detection. These findings fit the observation of original Taskononmy analysis~\citep{DBLP:conf/cvpr/ZamirSSGMS18}. 
As stated in the limitation, we leave the deeper analysis in Tasknonmy about spurious feature as future work, as currently we don't have the ground-truth anotation about which part of image input is causally related to each task.

\section{Detailed Hyper-parameter Selection Procedure and Sensitivity Analysis} \label{sec:hyper}

\begin{figure}[t!]
\begin{floatrow}
\ffigbox{%
  \includegraphics[width=1\columnwidth]{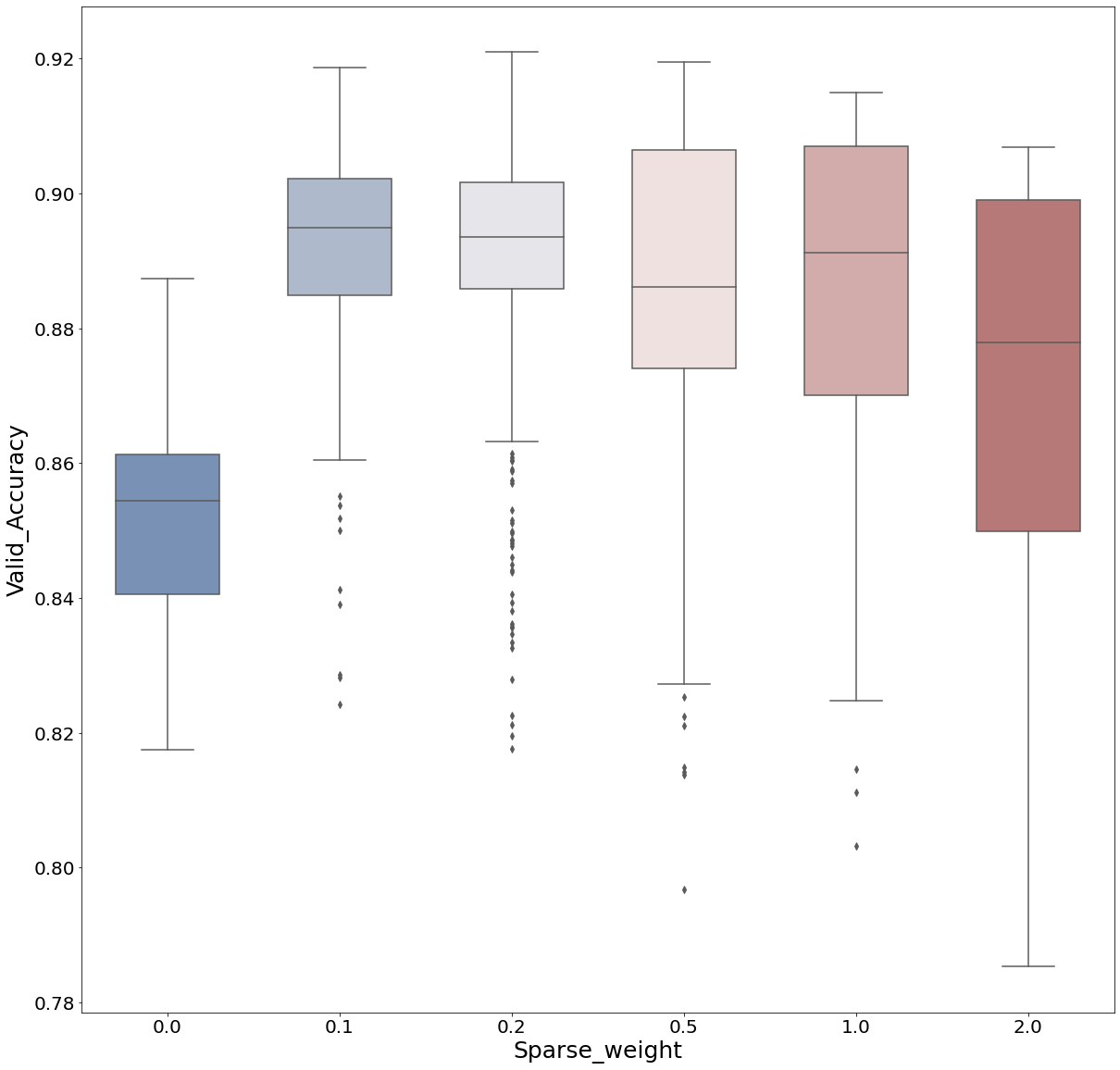}
}{%
  \caption{Hyper-parameter tuning results for sparse weight  ($\lambda_{sps}$) on Multi-MNIST.}
    \label{fig:hyper_1}
}
\ffigbox{
    \includegraphics[width=1\columnwidth]{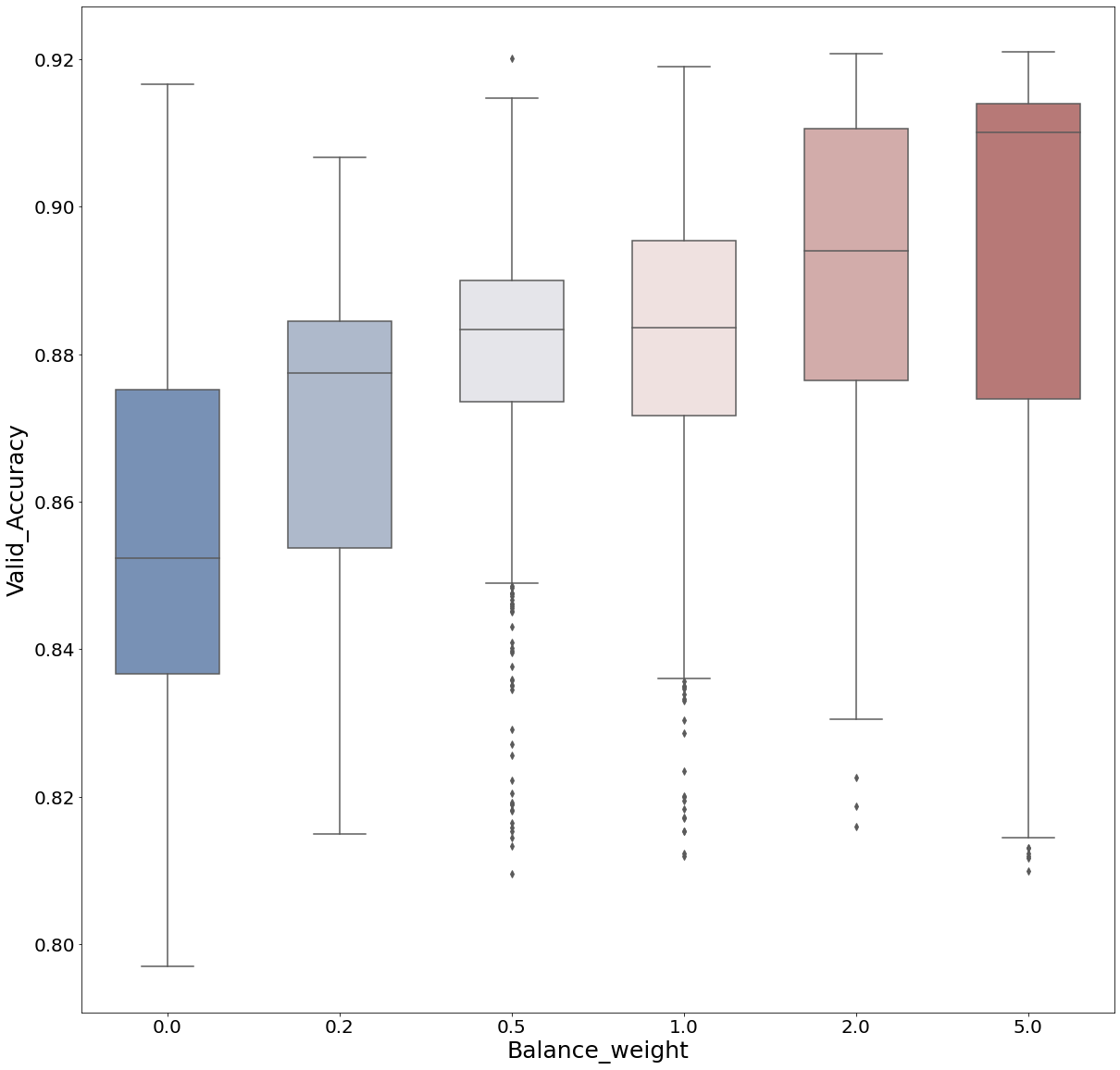}
}{%
   \caption{Hyper-parameter tuning results for balancing weight  ($\lambda_{bal}$) on Multi-MNIST.}
    \label{fig:hyper_2}
}
\end{floatrow}
\end{figure}

\begin{figure}[t!]
\begin{floatrow}
\ffigbox{%
  \includegraphics[width=1\columnwidth]{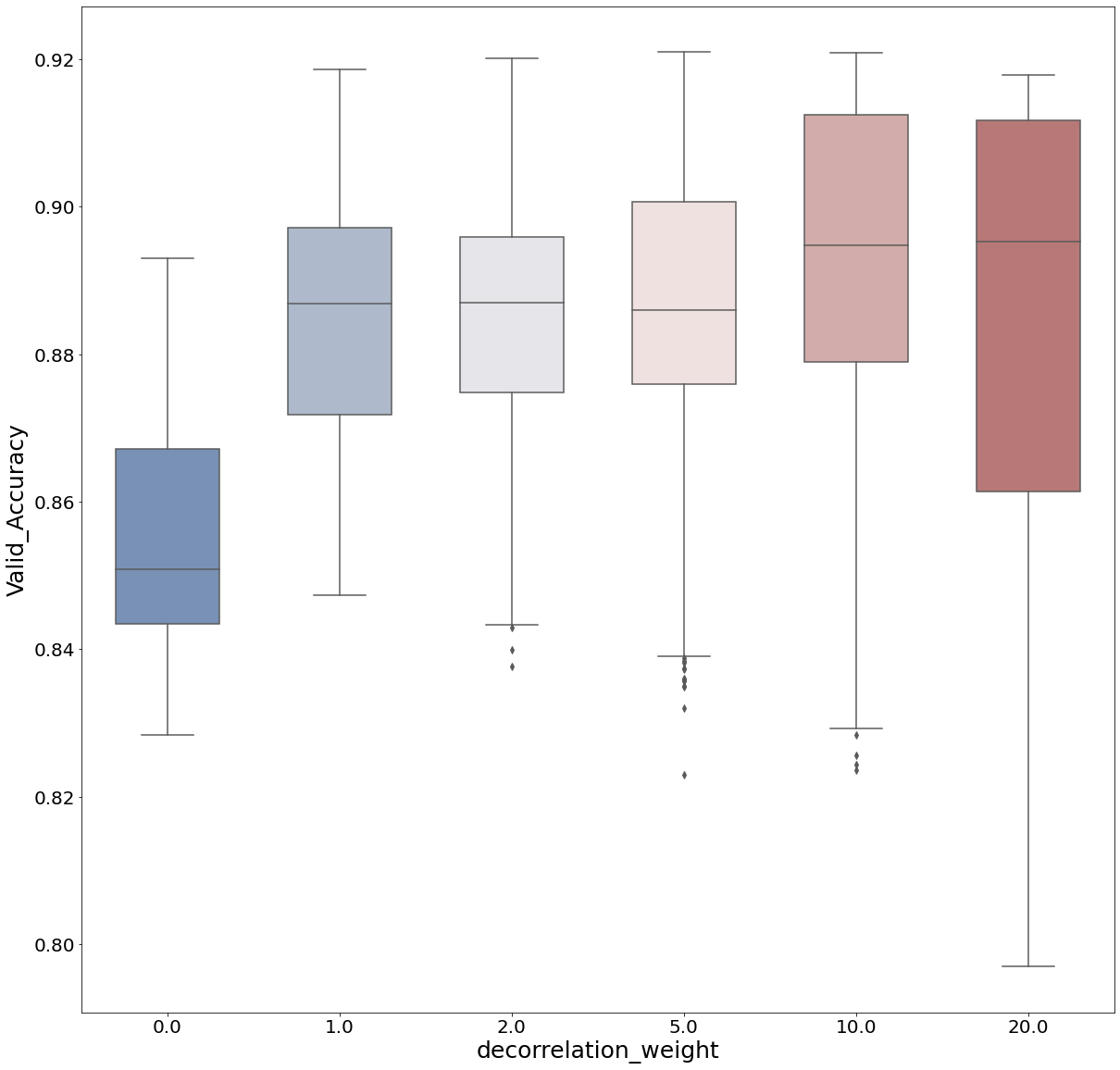}
}{%
  \caption{Hyper-parameter tuning results for disentanglement weight  ($\lambda_{decor}$) on Multi-MNIST.}
    \label{fig:hyper_3}
}
\ffigbox{
    \includegraphics[width=1\columnwidth]{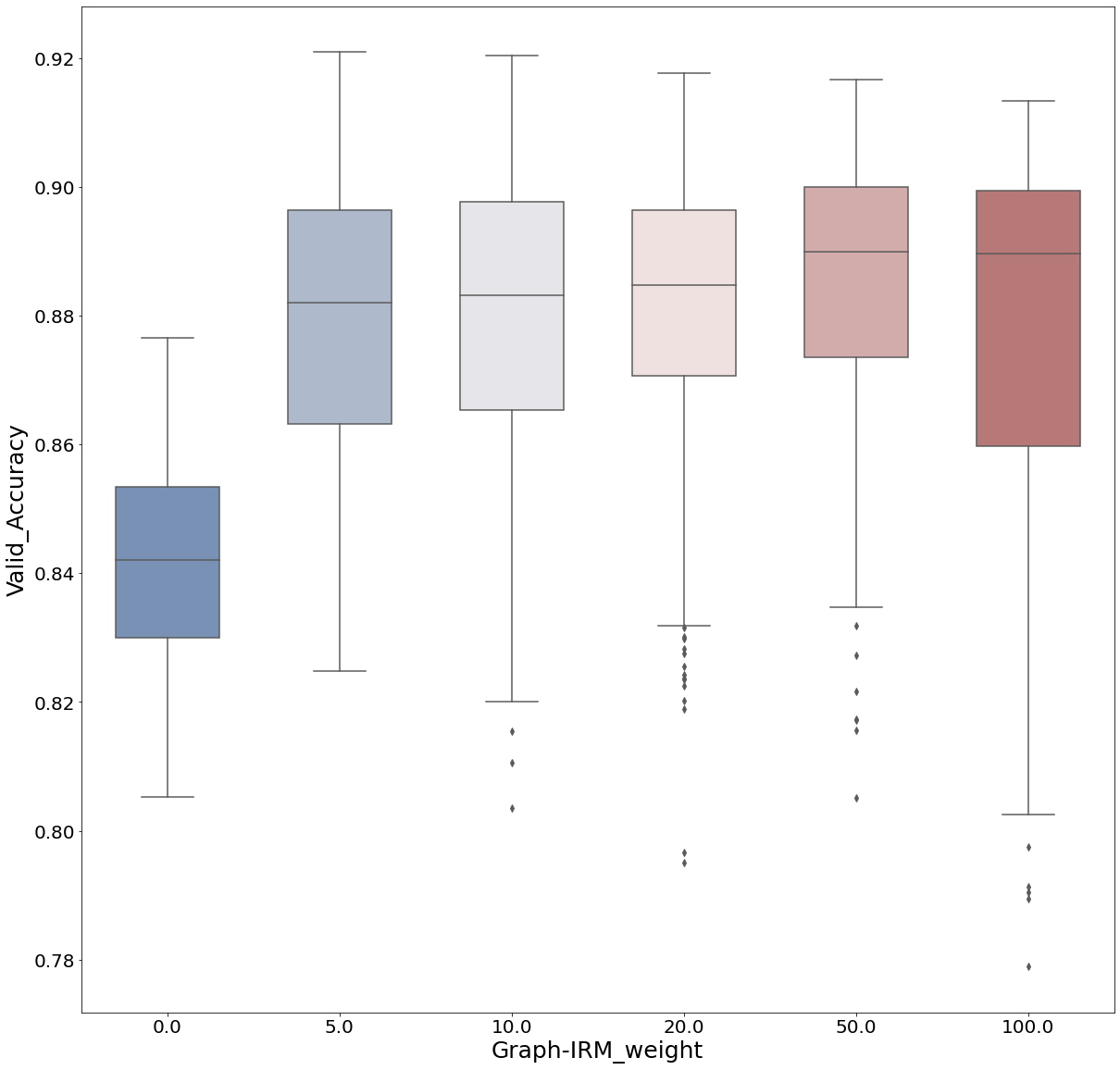}
}{%
   \caption{Hyper-parameter tuning results for correlation weight  ($\lambda_{G\text{-}IRM^{Var}}$) on Multi-MNIST.}
    \label{fig:hyper_4}
}
\end{floatrow}
\end{figure}

Here we introduce the procedure and results of hyper-parameter tuning.

Before discussing hyper-parameter selection, let me explain our baseline setup and experiment setting again. We have a single validation set, potentially bringing OOD to the training set. Our method only uses the training set to calculate the loss to update both encoder and per-task predictors. We then use the hold-out validation set to calculate the loss w.r.t graph weights A, and update it via invariance regularization. This avoids overfitting the validation set.

We mainly split the hyper-parameters into two sets: 
\begin{itemize}
    \item General hyper-parameter related to all baselines (including ours), including number of hidden states, optimizer, learning rate, number of modules ($K$);
    \item Our MT-CRL specific hyper-parameter, including weights for disentanglement ($\lambda_{decor}$), sparsity ($\lambda_{sps}$), balance ($\lambda_{bal}$), and invariance ($\lambda_{G-IRM}$).
\end{itemize}

For both sets of hyper-parameters, we only tune on the same OOD validation set we used for our method. 
All hyper-parameters except $K$ are standard hyperparameters for the MTL model. For datasets CityScape, NYUv2, and Taskonomy, we directly use the reported hyperparameter and dataset setting in previous papers~\citep{DBLP:conf/nips/YuK0LHF20, DBLP:journals/corr/abs-2010-02418}, in order to achieve a fair comparison. For MultiMNIST and MovieLens, we conduct a grid search for basic parameters, including the number of layers, number of hidden dimensions, optimizer, and learning rate, on the Vanilla MMoE MTL model without regularization. For the number of module ($K$), please refer to Sec~\ref{hyper}. After we determine these general hyper-parameters, we fix them and use them for all different MTL methods. This makes the comparison fair and ensures our performance improvement is not due to extensive hyper-parameter tuning of our method.

Next, we tune the MT-CRL-specific hyper-parameters on the validation set. We think this setting is reasonable as we didn't utilize this validation set to calculate training loss. Thus it could still be regarded as a whole-out set for most model parameters (except the graph weights, which only take a tiny portion of the whole model). Note that four regularization weight terms exist to be tuned, which is many burdens for the model. Therefore, we only use Multi-MNIST, the smallest dataset in all our testbeds, to conduct hyperparameter tuning for the ML-CRL-specific hyperparameters with grid-search. This is definitely not the best choice, and tuning for each dataset could potentially improve our performance further, but that only makes our improvement higher while not changing the main conclusion of this paper. Specifically, we choose the several ranges for the four regularization weights:
\begin{itemize}
    \item Sparse weight  ($\lambda_{sps}$): [$0.0, 0.1, 0.2, 0.5, 1.0, 2.0$]
    \item Balancing weight  ($\lambda_{bal}$): [$0.0, 0.2, 0.5, 1.0, 2.0, 5.0$]
    \item Disentanglement weight  ($\lambda_{decor}$): [$0.0, 1.0, 2.0, 5.0, 10.0, 20.0$]
    \item Invariance weight  ($\lambda_{G-IRM^{Var}}$): [$0.0, 5.0, 10.0, 20.0, 50.0, 100.0$]
\end{itemize}

These ranges are selected by running a few samples to determine the maximum value that should be within this range, and we keep each selection list to be a length of 6.
We report the boxplot of detailed results for each regularization weights in Figure~(\ref{fig:hyper_1}-\ref{fig:hyper_4}). As is illustrated, for all the regularization, using it is better than not using it ($\lambda=0$), showing their advantage in making our MT-CRL pipeline works. We then select the optimal hyperparameter that achieves the highest validation accuracy, which is $\lambda_{sps}=0.2,\lambda_{bal}=5.0, \lambda_{decor}=20, \lambda_{G-IRM}=5.0$. Again, this selection might not be the optimal solution; for example, the tendency for $\lambda_{bal}$ seems to increase with higher, and $\lambda_{G-IRM}$ might have a better choice within the range $[0-5]$. However, we did not conduct more searching and used this setup. After getting such a combination of MT-CRL-specific hyper-parameter, we fix it and use it for all other larger datasets, which assume our framework with this hyper-parameter combination is consistently effective. Further tuning them on a dedicated dataset should potentially bring better performance, but we did not do it to avoid the performance improvement brought by extensive tuning.

\paragraph{Sensitivity Analysis.}

From the curve and also the definition of these regularization, we know that for all other terms except sparsity regularization $\lambda_{sps}$, increasing the regularization weight and strictly force model to be balance, de-correlated or invariant doesn't harm too much to the model training (trend didn't go down even with relatively large weight). The only exception is the sparsity regularization. With high $\lambda_{sps}$ implemented as $L_1$ loss over adjacency weights will force all to be zero, which is very harmful to model training, which is why by default, we choose the value as $\lambda_{sps}=0.2$.

To give a simple example of whether our model is sensitive to an inappropriate setting of hyper-parameter, we run experiment on MultiMNIST and MovieLens, with the following randomly chosen hyper-parameter setting: $\lambda_{sps}=2.0,\lambda_{bal}=1.0, \lambda_{decor}=2.0, \lambda_{G-IRM^{Var}}=100.0$. The results compared with the original results are as shown in Table~\ref{tab:show}.

Note that with a randomly chosen hyper-parameter, the results on the two datasets drop slightly but are still significantly higher than the Vanilla MTL baseline. This is an informal showcase of our method's generality and not very sensitive to hyper-parameter selection.

\begin{table*}[t!]
\footnotesize
\centering
\begin{tabular}{l|cccccc} \toprule
 \textbf{Methods}    & \textbf{Multi-MNIST Accuracy} & \textbf{MovieLens MSE}  \\ \midrule
MT-CRL with $\mathcal{L}_{G\text{-}IRM}^{Var}$ and \textbf{default} hyper-parameter & 0.915 $\pm$ 0.018 &  0.884 $\pm$ 0.006 \\ 
MT-CRL with $\mathcal{L}_{G\text{-}IRM}^{Var}$  and \textbf{randomly chosen} hyper-parameter & 0.904 $\pm$ 0.021 &  0.887 $\pm$ 0.006 \\ 
Vanilla MTL baseline   & 0.846 $\pm$ 0.018 &  0.892 $\pm$ 0.005\\
\bottomrule
\end{tabular}
\caption{Results on Multi-MNIST and MovieLens with a randomly chosen set of hyper-parameter.}
\label{tab:show}
\end{table*}

\end{document}